\documentclass[lettersize,journal]{IEEEtran}

\usepackage{amssymb}
\usepackage{multirow}
\usepackage{makecell}
\usepackage[nointegrals]{wasysym}
\usepackage{bigdelim}
\usepackage{xspace}
\usepackage{comment}
\usepackage{graphicx}
\usepackage[hyphens]{url}
\usepackage{hyperref}
\usepackage{enumitem}

\usepackage{setspace}
\usepackage{amsfonts}
\usepackage{amsmath}
\usepackage{algorithmic}
\usepackage{textcomp}
\usepackage{xcolor}
\usepackage{enumitem}
\usepackage{verbatim}
\usepackage{booktabs}
\usepackage{footmisc}

\newcommand{\whitec}{\Circle}
\newcommand{\blackc}{\CIRCLE}
\newcommand{\grayc}{\LEFTcircle}

\newcommand{\reward}{\ensuremath{r}\xspace}
\newcommand{\return}{\ensuremath{\mathcal{R}}\xspace}
\newcommand{\policy}{\ensuremath{\pi\xspace}}
\newcommand{\stateRL}{\ensuremath{\vct s}\xspace}
\newcommand{\environment}{\ensuremath{\mathcal{E}}\xspace}
\newcommand{\valueRL}{\ensuremath{\mathcal{V}}\xspace}
\newcommand{\action}{\ensuremath{a}\xspace}
\newcommand{\model}{\ensuremath{\mathcal{M}}\xspace}
\newcommand{\observation}{\ensuremath{\vct o}\xspace}
\newcommand{\monitor}[1]{\ensuremath{m({#1})}\xspace}
\newcommand{\change}[1]{\ensuremath{c({#1})}\xspace}

\newcommand{\minimize}[1]{\ensuremath{\min ({#1})}\xspace}
\newcommand{\alter}[1]{\ensuremath{\text{learn}({#1}^U)}\xspace}
\newcommand{\reachTarget}[1]{\ensuremath{\text{reach}({#1}^T)}\xspace}
\newcommand{\learnTarget}[1]{\ensuremath{\text{learn}({#1}^T)}\xspace}
\newcommand{\learnTargetWhenState}[1]{\ensuremath{\text{learn}({#1}_{\vct s}^T)}\xspace}
\newcommand{\learnBackdoored}[1]{\ensuremath{\text{learn}({#1}^B)}\xspace}
\newcommand{\steal}[1]{\ensuremath{\text{steal}({#1})}\xspace}
\newcommand{\regularization}{\emph{reg.}\xspace}
\newcommand{\advtraining}{\emph{adv. tr.}\xspace}
\newcommand{\gametheory}{\emph{game th.}\xspace}

\newcommand{\distillation}{\emph{dist.}\xspace}
\newcommand{\detection}{\emph{det.}\xspace}
\newcommand{\sanitization}{\emph{san.}\xspace}
\newcommand{\ensemble}{\emph{ens.}\xspace}
\newcommand{\noAgent}{\ensuremath{\emptyset}\xspace}
\newcommand{\surrogate}{\ensuremath{\approx \model}\xspace}
\newcommand{\extended}{\ensuremath{\notin \environment}\xspace}
\newcommand{\within}{\ensuremath{\in \environment}\xspace}

\newcommand{\vct}[1]{\ensuremath{\mathbf{#1}}}

\newcommand{\set}[1]{\ensuremath{\mathcal{#1}}}

\newcommand{\myparagraph}[1]{\smallskip \noindent \textbf{#1}}

\newcommand{\ie}{{i.e.}\xspace}
\newcommand{\eg}{{e.g.}\xspace}
\newcommand{\etal}{{et al.}\xspace}

\usepackage{tikz}

\newcommand\submittedtext{%
  \footnotesize This work has been accepted for publication in IEEE Transactions on Intelligent Transportation Systems (T-ITS). The final published version will be available via IEEE Xplore.}

\newcommand\submittednotice{%
\begin{tikzpicture}[remember picture,overlay]
\node[anchor=south,yshift=10pt] at (current page.south) {\fbox{\parbox{\dimexpr0.65\textwidth-\fboxsep-\fboxrule\relax}{\submittedtext}}};
\end{tikzpicture}%
}

\begin{document}

\title{Security of Deep Reinforcement Learning for Autonomous Driving: A Survey}

%

\author{
    \IEEEauthorblockN{
    Ambra Demontis\IEEEauthorrefmark{1}, 
    Srishti Gupta\IEEEauthorrefmark{1}\IEEEauthorrefmark{2},  
    Maura Pintor\IEEEauthorrefmark{1}, Luca Demetrio\IEEEauthorrefmark{3}, Kathrin Grosse\IEEEauthorrefmark{4},
    Hsiao-Ying Lin\IEEEauthorrefmark{5},
    Chengfang Fang \IEEEauthorrefmark{6},
    Battista Biggio\IEEEauthorrefmark{1},
    Fabio Roli\IEEEauthorrefmark{3},
    }\\
    \IEEEauthorblockA{\IEEEauthorrefmark{1}University of Cagliari
    \{ambra.demontis, srishti.gupta, maura.pintor, battista.biggio\}@unica.it}\\
    \IEEEauthorblockA{\IEEEauthorrefmark{2}Sapienza, University of Rome
    \{srishti.gupta\}@uniroma1.it}\\
    \IEEEauthorblockA{\IEEEauthorrefmark{3}University of Genoa
    \{luca.demetrio, fabio.roli\}@unige.it}\\
    
    \IEEEauthorblockA{\IEEEauthorrefmark{4}IBM Research Zurich
    \{kathrin.grosse1\}@ibm.com}\\
    \IEEEauthorblockA{\IEEEauthorrefmark{5}Huawei Technologies France
    \{lin.hsiao.ying\}@huawei.com}\\
    \IEEEauthorblockA{\IEEEauthorrefmark{6}Huawei International
    \{fang.chengfang\}@huawei.com}
}

\maketitle

\submittednotice

\begin{abstract}
Reinforcement learning (RL) enables agents to learn optimal behaviors through interactions with their environment and has been increasingly deployed in safety-critical applications, including autonomous driving. Despite its promise, RL is susceptible to attacks designed either to compromise policy learning or to induce erroneous decisions by trained agents. Although the literature on RL security has grown rapidly and several surveys exist, these categorizations often fall short in guiding the selection of appropriate defenses for specific systems. In this work, we present a comprehensive survey of 86 recent studies on RL security, addressing these limitations by systematically categorizing attacks and defenses according to defined threat models for single- versus multi-agent settings. Furthermore, we examine the relevance and applicability of state-of-the-art attacks and defense mechanisms within the context of autonomous driving, providing insights that inform the design of robust RL systems.

\end{abstract}


\section{Introduction}

RL is a process in which machines learn to perform a task using a reward-based system. The machine tries to perform an action, and if it gets a positive (negative) reward/outcome, it will be more (less) likely to repeat that action in the future. 
For its ability to solve complex, dynamic and high-dimensional problems~\cite{nguyen21-tnnls}, RL is used in many applications. It is used even in high-risk high-value tasks such as cybersecurity-related ones~\cite{nguyen21-tnnls} and autonomous driving (AD)~\cite{kiran21-tits}. The former use RL to detect and fight against sophisticated cyberattacks, including, for example 
intrusion to host computers or networks~\cite{lopez-martin20-esa}, or malware \cite{wan17-globecom}. The latter extensively uses RL to perform tasks such as path planning and develop high-level driving policies for complex navigation~\cite{kiran21-tits}. 

Unfortunately, RL is highly vulnerable to adversarial attacks~\cite{behzadan18-arxiv, chen2019adversarial}
which can significantly alter agent behavior. In safety-critical domains such as AD, this may lead to severe consequences, including crashes or traffic disruptions, making RL security a pressing concern and an active research area. However, existing surveys remain limited: some review only early work~\cite{behzadan18-arxiv, chen2019adversarial}, others focus solely on attacks without defenses~\cite{shen2022sok}, or provide generic ML security overviews~\cite{shen2022sok, ibrahum_deep_2024, badjie25}. The authors of~\cite{standen_sok_2023} address a narrow subset of multi-agent systems. Consequently, a comprehensive overview of RL security, covering threat models, attack capabilities, defense shortcomings, and their alignment, remains absent. Moreover, current taxonomies~\cite{ilahi21-tai, lei_new_2023, mo24-CS} fail to guide system designers in selecting defenses tailored to specific RL models and attack scenarios. We summarize the shortcomings of existing work in Table~\ref{table-soa}. 

\begin{table}[t]
\caption{Limitations of existing work in DRL security.}\label{table-soa}
\centering
    \resizebox{0.99\linewidth}{!}{%

\begin{tabular}{r|c|c|c|c|c}
\toprule
                                     & Literature & Defenses      & RL-specific           & SARL/ & Defense \\
                                     & after 2022 &      &            &  MARL & guidance \\
\toprule
Behzadan et al.~\cite{behzadan18-arxiv}      &  -                                & \checkmark & \checkmark & \checkmark                 & -                                            \\
Chen et al.~\cite{chen2019adversarial}   & -                                & \checkmark & \checkmark & \checkmark                 & \checkmark                    \\
Shen et al.\cite{shen2022sok}           & \checkmark         & \checkmark & -                         & -                                         &-                                             \\
Ibrahum et al.~\cite{ibrahum_deep_2024}   & \checkmark         & \checkmark &  -                        & \checkmark                 & -                                            \\
Badjie et al.\cite{badjie25}              & \checkmark         & \checkmark & -                         &-                                          & \checkmark                    \\
Standen et al.~\cite{standen_sok_2023}    & \checkmark         & \checkmark & \checkmark & -                                         & \checkmark                    \\
Lei et al.~\cite{lei_new_2023}        & \checkmark         & \checkmark & \checkmark & \checkmark                 & -                                            \\
Mo et al.~\cite{mo24-CS}               & \checkmark         & \checkmark & \checkmark & \checkmark                 &  -                                           \\
Ours                                           & \checkmark         & \checkmark & \checkmark & \checkmark                 & \checkmark            \\
\toprule
\end{tabular}
}
\end{table}

Our survey aims to overcome the shortcomings mentioned above. The first contribution provided by our survey is a categorization that allows system designers to understand which defense they can apply to defend the system at hand against a precise threat. 
The second is to categorize all the literature on attacks and defenses against RL (more than 50 papers) accordingly.
Motivated by the fact that AD is a safety-critical application and companies such as Waymo and Tesla \cite{Jebessa22} are employing it in their AD cars, our third contribution is to provide a further application-specific taxonomy of the attack and defenses proposed for this application and to examine to which extent the attacks and defenses proposed against deep RL (DRL) can be applied to autonomous-driving systems. 
Unlike typical RL benchmarks such as games or robot control, autonomous vehicles operate in dynamic environments and open-world conditions, where decisions can directly affect real-world physical systems and public safety.
The environment of AD is characterized by partial observability, real-time constraints, and it is often affected by external systems or human drivers.
As a result, inspecting security aspects is crucial to prevent damage, harm, and traffic disruptions.
For this reason, assessing the security of DRL-based AD systems requires domain-specific threat models and robustness considerations that extend beyond the existing considerations found in the literature.
Finally, we explain that attacks and defenses against DRL are strongly inspired by those devised many years ago against machine learning classifiers~\cite{biggio18}. However, DRL has peculiarities that make the application of these attacks not straightforward. Therefore, only a subset of the attacks previously proposed against ML classifiers has been tested against DRL. This parallel allows us to provide our fourth contribution: a discussion about open challenges and research directions about security of DRL.

In Sec.\ref{sec:rl}, we introduce the RL background necessary to understand its vulnerabilities, followed by Sec.\ref{sec:RLforAD}, which outlines the specificities of RL algorithms in AD. Sec.\ref{sec:threat-model} presents a unified threat model for DRL security, while Sec.\ref{sec:attacks} and Sec.~\ref{sec:defenses} propose a novel taxonomy of attacks and defenses respectively. We then review existing works, assess their applicability to AD, and highlight open research directions.

To summarize, we provide the following contributions.
\begin{itemize}
\item A taxonomy of DRL attacks and defenses that enables specific threats with suitable protections;

\item A categorization of over 80 papers on DRL security;
\item 
An analysis of the applicability of state-of-the-art attacks and defenses specific to AD;

\item A comparison of DRL and ML security, highlighting open challenges and future research directions.
\end{itemize}

\section{Reinforcement Learning}
\label{sec:rl}
DRL enables agents to learn from experience repeating rewarding actions and avoiding those with negative outcomes. DRL has been successfully applied to complex, dynamic tasks such as AD, where vehicles must navigate traffic and anticipate the behavior of other road users, even under rule violations~\cite{kiran21-tits}.
In this section, we provide the necessary DRL background: first describing its core components, then outlining how state-of-the-art systems can be categorized.

\subsection{Components of Reinforcement Learning Systems}
We assume familiarity with the standard RL concepts~\cite{sutton18}, which we briefly summarize here for the purpose of contextualizing the survey. 
In the following, we describe the components of an RL system and their interactions. The notation used in this survey is summarized in Table~\ref{tab:notation}.

\begin{enumerate}[label=\alph*)]
\item\textit{Agent:} An agent is an entity that can take actions and influence its environment. In the AD scenario, each car on the street and each pedestrian can be an agent. The agent trained using RL is referred to as \emph{ego} agent. In this work, we only consider RL agents.
\item \textit{Action:} An action $\action$ is a move an agent can take at time $t$, denoted $\action_t$, chosen from the set of all possible actions: $\set A$. In AD, actions include turning, accelerating, decelerating, or maintaining course.

\item \textit{Environment:} The environment $\environment$ is the scenario in which agents operate, such as a street with cars and pedestrians. It also includes other relevant elements, such as traffic lights and signals.

\item \textit{State:} The state $\stateRL$ encodes the environment’s conditions, such as ego car positions or traffic light colors. Based on the agent's action, the environment states are updated: for instance, if the agent turns right at time $t$ ($\action_t$), the resulting state at $t+1$ is $\stateRL_{t+1}$.

\item \textit{Reward:} The reward $\reward$ measures the success of an agent’s action, provided by the environment. For example, the ego car receives a \emph{positive} reward for moving closer to its destination and a \emph{negative} reward for collisions. The reward for action $\action_t$ at time $t$ is denoted $\reward_t$.

\item \textit{Observations:} An observation $\observation$ is the information an agent receives from the environment at a given time step to make decisions. It may not fully reflect the true state, for example, due to sensor limitations.

\item \textit{Policy:} The policy $\policy$ is the strategy (or behavior) adopted by the agent to infer the best action to take at its state $\stateRL$. 

\end{enumerate}

\begin{table}[!ht]
    \caption{Summary of the notation and abbreviations.}\label{tab:notation}
    \centering
    \resizebox{0.5\textwidth}{!}{%
    \begin{tabular}{ll}
    \toprule
\multicolumn{1}{l}{\textbf{RL Symbols}}        &                                                                          \\
\action                                     & Action                                                                   \\
\environment                                & Environment                                                              \\
\stateRL                                    & State                                                                    \\
$\reward$                                                  & Reward                                                                   \\
\observation                                & Observations                                                             \\
\return                                     & Return                                                                   \\
$\gamma$                                                   & Discount factor                                                          \\
\policy                                     & Policy                                                                   \\
\valueRL                                    & Value function                                                           \\
\model                                      & Model                                                                    \\ \hline
\multicolumn{1}{l}{\textbf{Attack Violations}}                        &                                                                          \\
\change{x}                                    & The attacker changes some inputs $x$                \\
\monitor{x}                                   & The attacker monitors some inputs/outputs $x$               \\
\minimize\return             & Minimize the return                                                      \\
\reachTarget\stateRL         & Make \environment reach a target state $\vct s^T$         \\
\learnTarget\policy          & Make the agent learn a target policy                                     \\
\alter{\policy} &  Make the agent deviate from intented policy\\

\learnBackdoored\policy      & Make the agent learn a policy containing a backdoor                      \\
\learnTargetWhenState\policy & Make the agent learn to use a target policy  \\
& for a subset of observations\\
\steal\model                 & Attacker copies model \model without consent  \\        
\extended                                   & Attack uses agent that is not part of the environment     \\    
\within                                     & Attack's agent is within the environment                           \\
\surrogate                                  & Attack's agent is a surrogate (copy of the target) agent           \\
\hline
\multicolumn{1}{l}{\textbf{Defense Methods}}               &                                                                          \\
\detection                                  & Detection                                                                \\ 
\sanitization                               & Sanitization                                                             \\
\advtraining                                                         & Adversarial training                                                     \\
\gametheory                             & Game Theory                                                             \\
\distillation                               & Distillation                                                             \\
\regularization                             & Regularization                                                           \\
\ensemble                             & Ensemble                                                           \\
\bottomrule
\end{tabular}
}
\end{table}

\subsection{Solving the Reinforcement Learning Problem}
RL studies how an agent learns to act in an environment by maximizing cumulative reward through interaction. An RL problem is commonly modeled as a Markov Decision Process (MDP), defined by a tuple $(\set S, \set A, P, R, \gamma )$, where $\set S$ is the state space, $\set A$ is the action space, $P(\stateRL^\prime|\stateRL, \action)$ defines the transition dynamics, $\reward(\stateRL, \action)$ is the reward function that cumulates in $R$, and $\gamma \in [0, 1]$ is the discount factor.
At time step $t$, the agent observes a state $\stateRL_t$, selects an action $\action_t$, receives a reward $\reward_t$, and transitions to state $\stateRL_{t+1}$.
The objective is to maximize the expected discounted return: $
R_t = \sum_{i=t}^{\infty} \gamma^{\, i-t} r_i$.
A policy $\policy(\action|\stateRL)$ defines the agent's behavior. 
The state-value and action-value functions are:
\begin{equation}
V^{\policy}(\stateRL) = \mathbb{E}_{\pi}[R_t \mid \stateRL_t = \stateRL], \,
Q^{\policy}(\stateRL,\action) = \mathbb{E}_{\pi}[R_t \mid \stateRL_t = \stateRL, \action_t = \action].
\end{equation}
RL methods learn an optimal policy $\policy^*$ either by estimating $V$, learning the policy directly, or combining both.

\subsection{Categorization of Reinforcement Learning Systems}\label{sec:categorization}

In this section, we explain how state-of-the-art RL algorithms can be classified based on specific features, followed by a discussion on exemplary approaches for each category.

\myparagraph{Single-Agent vs Multi-Agent RL.}  RL environments can incorporate a \emph{single-agent} (SARL) or \emph{multi-agents} (MARL). In SARL, an ego agent learns a policy to interact with the environment. In MARL, multiple RL agents, each with its own policy and objectives, interact with a shared environment and potentially with each other. This makes the system more complex as agents’ actions influence the state and rewards of others. MARL rewards can be cooperative (shared goal), competitive (opposing goals), or mixed-sum (combining both). MARL often models cooperative behaviors, such as minimizing traffic congestion or avoiding collisions~\cite{wang2020stop}, though mixed-sum setups may apply in special cases, \eg, when regular vehicles must yield to ambulances while still pursuing their own objectives. 
In this work, we consider only systems with RL-based agents as MARL; other entities, such as pedestrians or human-driven cars, are not considered agents.

\myparagraph{Model-based vs Model-free RL.} RL can be model-based or model-free~\cite{kaelbling1996reinforcement}. In \emph{model-based} RL, the agent learns or simulates an explicit model of the environment’s dynamics, denoted \model, including a transition function (how actions change states) and a reward function (expected rewards for state-action pairs), enabling planning and foresight. 
In contrast, \emph{model-free} RL learns directly from interactions with the environment, without modeling its dynamics explicitly. 
Model-based RL allows planning but incurs higher sample complexity, cost, and risk of inaccuracies.  

\myparagraph{On-policy vs Off-policy vs Offline RL} Depending on \emph{how} the RL agent learns a policy from its environment during the training phase, the RL methods can be classified as a) on-policy, b) off-policy, and c) offline RL methods. In \emph{on-policy} methods, the target policy being learned is the same as the behavior policy used to collect data, simultaneously exploring and updating the policy, as in PPO~\cite{gleave2019adversarial}.
\emph{Off-policy} methods learn the value of a target policy using data collected by a different behavior policy, allowing reuse of past interactions, as in DQN~\cite{zhao2020blackbox}. 
\emph{Offline} RL learns entirely from pre-collected static datasets without environment interaction~\cite{poole2010artificial}. Most RL algorithms in this survey are on- or off-policy, except the planning-based offline agent studied by~\cite{rakhsha2020policy}.

\section{Reinforcement Learning for Autonomous Driving}\label{sec:RLforAD}

One widely used application area for DRL is AD. 
We first link the DRL categorization to AD tasks (Sec.\ref{sec:adtasks}), then discuss car components to show how DRL is applied and interacts with them (Sec.\ref{sec:components}).

\subsection{Driving Automation and Reinforcement Learning}\label{sec:adtasks}
To relate the previous categorization to AD, we review the six Society of Automotive Engineers (SAE) automation levels~\cite{sae20143016}.
We will start with the lowest level and progress towards more autonomy. 
\begin{enumerate}
    \item[] \textbf{Level 0.} No automation, automatic systems may provide warnings or assistance. Examples: lane departure warnings, emergency braking.  
    \item[] \textbf{Level 1.} The driver controls the car, the system provides support by \emph{either} braking \emph{or} accelerating. Examples: lane centering, cruise control.
    \item[] \textbf{Level 2.} The driver controls the car, the system supports the driver by braking \emph{and} accelerating. Examples: simultaneous lane centering and cruise control.
    \item[] \textbf{Level 3.} The system controls the car, but the driver may be asked to take over control at any time. Examples: traffic jam chauffeur. 
    \item[] \textbf{Level 4.} The system controls the car without any human intervention, but it can only drive in specific areas. Examples: driverless taxi restricted to a certified road.
    \item[] \textbf{Level 5.} The system autonomously controls the functioning of the car without human intervention.
\end{enumerate}

Automation levels directly impact the RL application, particularly the agent’s action set. Higher levels (4–5) often involve multi-agent settings, where agents collaborate, e.g., exchanging position information for safe driving. Lower levels (0–3) focus on single-agent tasks, such as lane keeping, with limited interaction.
Environmental complexity also affects the choice between model-based and model-free RL. High automation increases environmental complexity, making explicit modeling difficult; thus, model-free approaches are typically favored.
Policy update strategy depends on safety considerations. Offline RL trains on pre-collected data to avoid unsafe actions, but can limit exploration outside the training distribution. However, on- and off-policy training can still be trained in virtual environments. In deployed vehicles, manufacturers may periodically update mature policies.\footnote{\url{https://www.tesla.com/support/software-updates}}

\subsection{Autonomous Driving Components}
\label{sec:components}

\begin{figure*}[t]
    \centering
    \includegraphics[width=0.85\textwidth]{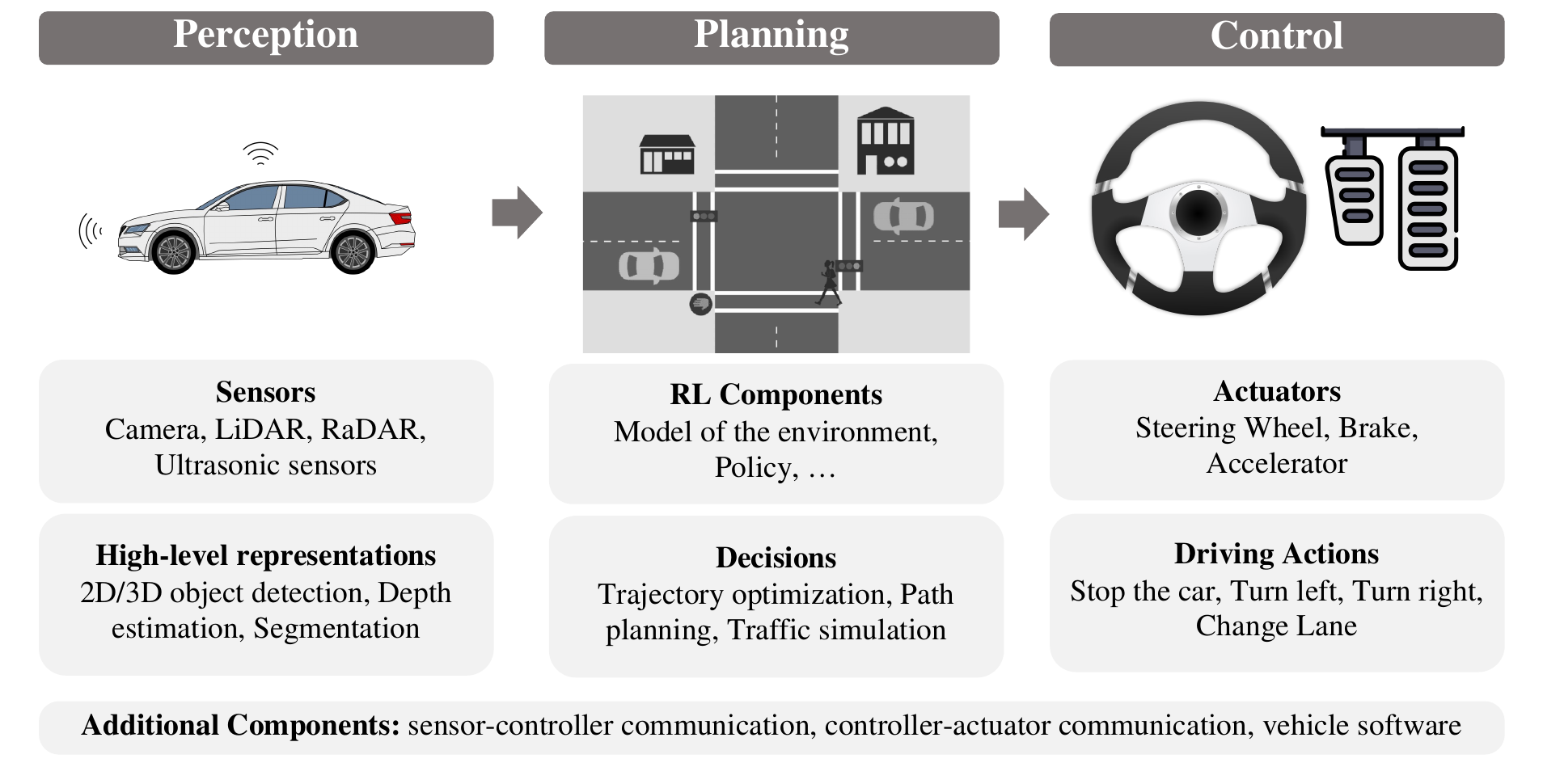}
    \caption{Major Components of Autonomous Driving Systems.}
    \label{fig:ad_components}
\end{figure*}

To interface with the environment, different types of sensors and components are required. Previous work describes these components at different levels of abstraction~\cite{garcia2020comprehensive,qayyum2020securing}.
We chose the three-level abstraction by Pendleton et al.~\cite{pendleton2017perception}, which divides the information flow through the vehicular agent into \emph{perception} \emph{planning}, and \emph{control}. 
We illustrate the main components of the abstraction in Figure~\ref{fig:ad_components}, and we describe them in the remainder of this section.

Some of the commonly used sensors~\cite{georgia2021cybersecurity} are: LiDAR, with a range of up to 200 meters often used for collision avoidance. Another sensor similar to LiDAR is Radar, with a similar range, and it may also be used for collision detection in the closer vicinity of the vehicle.
Both inputs can be represented, for example, as point clouds. Other sensors include cameras for object detection, which typically cover a range of up to 100 meters and produce images or videos.
Ultrasonic sensors have a short range and are often used for parking assistance. 
Several sensors are often used in conjunction for redundancy and thus increase the reliability of collected inputs. 
In addition, this layer also incorporates tasks such as object detection, depth estimation, and semantic segmentation, where inputs are already processed, resulting in aggregated information.

\myparagraph{Planning.} 
Receiving the previously collected inputs and inputs from other vehicles (in case of MARL), an RL agent can be trained to plan actions to achieve the final goal. Depending on the goal, the agent may need to perform \emph{route planning}: planning which roads to be taken to reach the destination, \emph{behavioral planning}: how to interact with other agents in the vicinity while following rules and restrictions, \emph{motion planning}: generate location objectives and determine actions to achieve them.
In the following sections, we will discuss tasks such as deriving the prediction of the steering output from input images~\cite{yang2020enhanced,boloor2019simple,sun2020stealthy,xiao2019characterizing}, traffic simulation~\cite{wang2020stop}, and path finding~\cite{ma2019policy}. 
Depending on the task, the requirements of an RL agent may vary greatly. In one case, an ego agent can maneuver a car on a highway alongside other non-autonomous vehicles. In this case, the RL system must be guaranteed to operate in \textit{real-time}. 
However, in another case, small delays may be more acceptable. For example, when an agent operates in an area void of humans; both pedestrians and drivers, or in MARL settings, where multiple agents can act simultaneously or take into account delays of other vehicles. 

\myparagraph{Control.} 
Once target actions have been identified, the agent needs to execute them to control vehicle movements. This step involves communicating signals to the actuators to perform the required actions. 
Depending on how the actions are encoded, they might be low-level actions, \ie, the output is directly the input to the actuators, or high-level actions, \eg, ``change lane'', which is then performed with a sequence of predefined low-level commands.

\myparagraph{Further Components.}
There are other parts of the autonomous vehicle that we skipped previously for simplicity. For example, the application of sensors requires software.  

Another example is the layers required to translate the RL agent's output into the autonomous vehicle's control. Each component may include a security vulnerability. Moreover, the vehicle may communicate with infrastructure, other vehicles, pedestrians, and roadside units.  There are also communications within the vehicle (sensor-controller communication e controller-actuator communication).
Some attacks assume that the attack does not alter the environment, but the information provided to the RL agent~\cite{zhao2020blackbox,sun2020stealthy,tretschk2018sequential}.

While the \emph{perception} components are vulnerable to specific adversarial attacks~\cite{shen2022sok} that target them independently of the underlying system, for the sake of this survey, we will focus on attacks targeting the RL component specifically. Similarly, for technical papers that address the security of \emph{control} components individually, we refer the reader to related work~\cite{bhamare2020cybersecurity}. 

\myparagraph{Toward End-to-End and Integrated Pipelines.}
More recently, AD has also shifted toward end-to-end learning~\cite{codevilla2018imitation,chitta2023transfuser}. In this setting, the RL model directly maps raw sensor data to decision-making and driving commands.
Consequently, the boundaries between perception, planning, and control become less evident, exposing a cross-module attack surface. We discuss attacks against end-to-end driving in Sect.~\ref{sec::ADAttacks}.

\subsection{Cooperative Driving System (CDS).} 
\label{sec::cds}

CDS refers to a situation in which multiple road actors like vehicles, infrastructure, etc actively coordinate their behavior to improve traffic flow. This is typically enabled through interaction-aware and communication-enabled agents. These agents exchange information such as position, velocity, intent, or environment perception. Typical driving scenarios include Platooning, where a group of vehicles travels closely to improve traffic flow; Cooperative Merging, \ie vehicles collaborate to allow smooth merging at highway ramps and lane closures; Intersection Management \ie when instead of traffic lights, vehicles negotiate crossing orders to reduce waiting times, etc. While cooperative driving offers traffic control benefits, it also open doors to new attack surfaces for malicious entities as we will see in~\autoref{sec::ADAttacks}.

\section{Threat Model}
\label{sec:threat-model}

In this section, we describe a threat model for RL. 
We characterize the attacks according to (1) the goal of the attacker, (2) the knowledge the attacker has about the target system, and (3) the attacker's capabilities of manipulating input data and/or tampering with the system components. 

\subsection{Goal}
The attacker's goal is the result they aim to achieve by perpetrating the attack. The goals can be broadly categorized according to the type of security violation they cause: a) \emph{integrity} violation, \ie, to influence the agent to perform an unintended action. For example, it can force an autonomous-vehicle agent to steer the car in the wrong direction when making a turn. b) \emph{availability} violation, \ie, when the legitimate functioning of the agent is compromised. For example, an attacker can alter a portion of the training data, making the agent unable to learn to perform certain tasks, and finally, c) \emph{privacy} violation by obtaining private information about the system or the data used to train it.

\subsection{Capabilities}
Attackers may have different capabilities that they can leverage to achieve their goals. 
Firstly, they may or may not have the ability to alter training data or the learning process. 
If the attacker can modify what the target agent learns, then the attack is defined as a training-time attack. 
Attackers can also alter the code of the RL algorithm that is used to train the agent, as done in~\cite{mandlekar2017adversarially}.
If the attacker can act only against an already trained agent, it is defined as a test-time attack. 

\subsection{Action}

\begin{figure}[t]
    \centering
    \includegraphics[width=\linewidth]{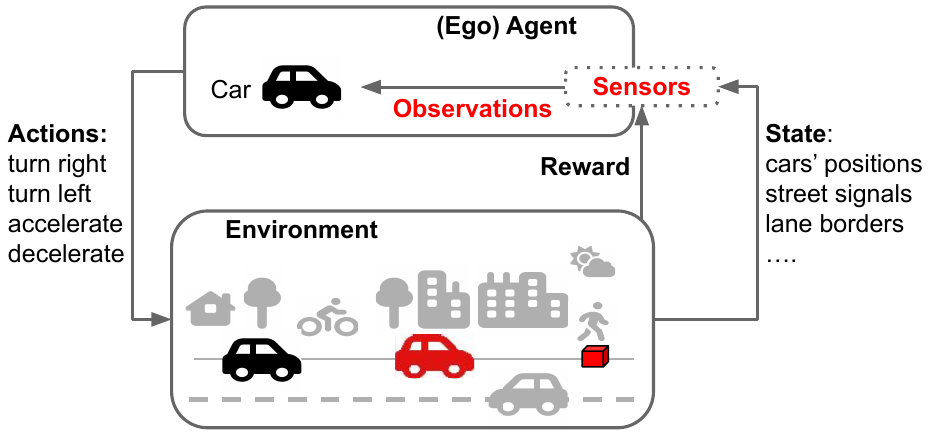}
    \caption{Conceptual representation of the components of an RL policy that the attacker can manipulate without having access to the RL algorithm. }
    \label{fig:RL-for-AD-attacks}
\end{figure}

Depending on the threat model, an attacker can either change (\change{x}) or monitor (\monitor{x}) the target system’s inputs and outputs. The attacker has higher flexibility when changing the input to perform a desired targeted attack. Since the agent’s observations drive both training and inference, attackers can perturb the environment (\change{\environment}) by adding objects or introducing malicious agents, alter specific states (\change{\stateRL}), or tamper with sensory inputs (\change{\observation}) to modify the agent’s perception. They can also manipulate actuators (\change{\action}) such that the action performed differs from those chosen by the agent, or modify the reward signal (\change{\reward}) during training to degrade performance. Depending on the system, attackers may combine multiple capabilities. In integrated end-to-end pipelines, the attacker may not need to explicitly target the RL module; compromising perception may indirectly manipulate planning and control.
Figure~\ref{fig:RL-for-AD-attacks} highlights, in red, the components that can be manipulated without access to the RL algorithm.

\subsection{Knowledge}
The attacker can have different levels of knowledge of the target system, including 
(i) the specifics of the model, \eg, the type of observations or actions, the DRL algorithm used, the reward function, (ii) the training data, (iii) the internal parameters or values which might include the models' weights or the learned policy, the values of the observations, actions, or reward at each step. If attackers have full knowledge of the system, the attack is termed a \emph{white-box}. Though unrealistic in practice, this setting enables worst-case scenario evaluations and provides upper bounds on performance degradation. In a \emph{gray-box} setting, adversaries have partial knowledge—for instance, they may observe states and rewards but lack details of the DRL algorithm. Some variants assume delayed knowledge acquisition, in which the system state changes before the attacker gains full access. Finally, in a \emph{black-box} setting, the attacker has no internal knowledge and relies solely on external observations of inputs and outputs, such as monitoring agent actions. In gray- and black-box settings, attackers often exploit transferability~\cite{demontis19-usenix}, crafting attacks on a surrogate and deploying them against the target.



\myparagraph{Leveraging Knowledge for Malicious Goals.} 
When running attacks against RL systems, attackers can leverage information about the environment to achieve their malicious goals. There are several ways in which this can happen:

\begin{itemize}
\item When attackers do not have perfect knowledge about their target system, they train a surrogate model to approximate the target model. We will denote the surrogate system by \surrogate.

\item In RL systems, multiple agents can act simultaneously in the environment and thus modify the state as perceived by the victim. There are two ways in which the attacker can deploy the attack in an RL environment:
\begin{itemize}
    \item The attacker can use an agent from within the environment (\within) to carry out the attack. 
    \item The attacker can also leverage an agent \textit{not} included in the environment (\extended) to enhance the severity of the attack. For example, when the attacker uses an external agent that sneaks into the victim agent's environment and later uses this information to deploy the attack.
\end{itemize}
\end{itemize}
In the following, we will use the symbol \noAgent to denote the cases in which the attacker does not use any malicious RL agent.

\section{Attacks}
\label{sec:attacks}
Despite its potential, DRL is vulnerable to numerous attacks, many of which are adapted from ML. Here, we categorize state-of-the-art RL attacks and present a taxonomy based on attacker capabilities. We also review studies in AD, highlighting assumptions and real-world applicability, and outline future directions for DRL-specific attack research.

\subsection{Categorizing Attacks against DRL}
\label{sec: attack_category}

Our taxonomy is designed to guide both researchers and practitioners by highlighting attack characteristics and the DRL systems they target. This helps readers identify applicable attacks and potential vulnerabilities before deployment. Attack goals depend on the target system, and factors such as policy update strategy and environment modeling influence attack feasibility. Accordingly, we first identify key characteristics of DRL models that are critical for understanding possible attacks.

\myparagraph{Agent Model.} Characteristics of the DRL agent under attack.
\begin{itemize}
\item \emph{Single vs. Multi Agent}: Methods are evaluated in either SARL or MARL settings, depending on the number of agents. MARL represents a more realistic but challenging scenario for AD.
\item \emph{Target Policy Update.} RL policy update frequency (on-policy, off-policy, offline). This determines whether training-time attacks can have an immediate impact.
\item \emph{Model-based.} Indicates whether the DRL system models the environment. If true, attacks can potentially target its ability to predict environment dynamics; otherwise, we mark it as false.
\end{itemize}
\myparagraph{Threat Model.} Capabilities of the attacker: 
\begin{itemize}
    \item \emph{Attack's Time.} Whether the attack is planted at training time or test time.
    \item \emph{Attacker's Action.} What action must be performed by the attacker, and, as we discussed, could be to monitor the inputs of the RL system or to alter them. 
    \item \emph{Poisoning.} Ability of the attackers to alter the input during training, which can be true if they have this capability and false otherwise.
    \item \emph{Attacker's Knowledge.} Level of knowledge of the victim system, which may be white-, gray-, and black-box.
    \item \emph{Attacker's goal.} The result that the attackers aim to obtain with their attack.
    \item \emph{Sequential Attack.} The eventual temporal dependence of the attack on the current or previous state. Sequential attacks are more complex to execute, as the attacker must know or predict the states that occur earlier or later than the state in which the attack occurs.
    \item \emph{Attacker's Agent.} Whether the attacker uses an RL agent to carry out the attack, either by deploying the agent inside or outside the RL environment, or by simply training a surrogate model.
\end{itemize}

{\renewcommand{\arraystretch}{1.3}   

\begin{table}
\centering
\caption{The attacks against RL, categorized according to the attackers' capability and the security violation they cause.}
\label{tab:attacks-categorization}
\resizebox{0.45\textwidth}{!}{%
\begin{tabular}{c|l|l|l|}
\multicolumn{1}{l}{}           & \multicolumn{1}{c}{Integrity} & \multicolumn{1}{c}{Availability} & \multicolumn{1}{c}{Privacy}  \\ 
\cline{2-4}
\multirow{2}{*}{Test Data} & - \reachTarget {\stateRL}            &                                  & -  \steal {\model }               \\
                               & - \minimize {\return}       &                                  &                              \\
\cline{2-4}
\multirow{2}{*}{Training Data}
                               & - \learnBackdoored {\policy }     & - \learnTarget {\policy}           &                              \\
                               & - \learnTargetWhenState{\policy}                              & - \alter{\policy}               &                              \\
\cline{2-4}
\end{tabular}
}
\end{table}}

Table~\ref{tab:attacks-categorization} categorizes attacks by the type of security violation: integrity, availability, or privacy, caused at train or test time, described further in detail in the next two sections. Additionally, Table\ref{tab:Attacks} provides a schematic overview of 50 recently proposed attacks. Very few attacks target model-based systems, and none of them are world-based. Given that a single interaction with the environment can have a large impact on world-based systems, we expect poisoning attacks to be more effective on them than against other model-based systems.  



\begin{table*}[!ht]
 \caption{Attacks against reinforcement learning systems. The presence of the \checkmark indicates that the corresponding attribute is true for the attack. For the attacker’s knowledge, we use \blackc \ to represent black-box,  \grayc \ for gray-box, and \whitec \ for white-box. Refer Table\ref{tab:notation} for attacker's agent in the system and Table\ref{tab:attacks-categorization} for attacker's goal. 
 }
 \label{tab:Attacks}
    \centering
    \resizebox{0.99\textwidth}{!}{%
    \begin{tabular}{r|rrrrrrrrrr}
    \toprule
       & \multicolumn{3}{c}{\textbf{AGENT MODEL}} & \multicolumn{7}{c}{\textbf{THREAT MODEL}}
        \\
        \cmidrule(l){2-4} \cmidrule(l){5-11}
       References &
       
       Single vs. &
       Policy &
       Model &
       Attack's &
       Attacker's &
       Poisoning &
       Attacker's &
       Attacker's &
       Sequential &
       Attacker's \\
        & 
        
        Multi-agent &
        Update &
        Based &
        Time &
        Action &
        &
       Knowledge &
       Goal &
       Attack &
       Agent \\
        \midrule
        Wang \etal \cite{wang2020stop} & MARL & off-policy &  & $tr$ & \change{r} & \checkmark & \blackc & \learnBackdoored{\policy} &  & \within \\
        Gleave \etal \cite{gleave2019adversarial} & MARL & on-policy &  & $ts$ & \change{\environment} &  & \grayc & \minimize{\return} & \checkmark & \within \\ 
        Zhao \etal \cite{zhao2020blackbox} & SARL & on/off-policy &  & $ts$ & \change{\observation} &  & \blackc & \minimize{\return} &  & \noAgent \\
        Rakhsha \etal \cite{rakhsha2020policy} & SARL & on-policy, offline & \checkmark & $tr$ & \change{r,\action,\stateRL} & \checkmark & \whitec & \learnTarget{\policy} & \checkmark & \noAgent \\ 
        Yang \etal \cite{yang2020enhanced} & SARL & off-policy &  & $ts$ & \change{\observation} &  & \whitec & \minimize{\return} & \checkmark & \extended \\ 
        Yang \etal \cite{yang2020enhanced} & SARL & off-policy &  & $ts$ & \change{\observation} & ~ & \blackc & \minimize{\return} & \checkmark & \surrogate \\ 
         Boloor \etal \cite{boloor2019simple} & SARL & on-policy &  & $ts$ & \change{\environment} &  & \whitec & \reachTarget{\stateRL} & \checkmark & \noAgent \\ 
         Sun \etal \cite{sun2020stealthy} & SARL & on/off-policy &  & $ts$ & \change{\observation} &  & \whitec & \minimize{\return} & \checkmark & \extended \\ 
        Sun \etal \cite{sun2020stealthy} & SARL & on/off-policy &  & $ts$ & \change{\observation} &  & \grayc & \minimize{\return} & \checkmark & \extended \\ 
         Xiao \etal \cite{xiao2019characterizing} & SARL & off-policy &  & $ts$ & \change{\observation} &  & \whitec / \blackc & \minimize{\return} & \checkmark & \noAgent \\ 
        Xiao \etal \cite{xiao2019characterizing} & SARL & off-policy &  & $ts$ & \change{\action} &  & \whitec & \minimize{\return} &  & \surrogate \\ 
        Xiao \etal \cite{xiao2019characterizing} & SARL & off-policy &  & $ts$ & \change{\environment} &  & \blackc & \reachTarget{\stateRL} & N/A & \noAgent \\ 
         Ma \etal\cite{ma2019policy} & SARL & on-policy &  & $tr$ & \change{r} & \checkmark & \whitec & \learnTarget{\policy} &  & \surrogate \\ 
         Tretschk \etal \cite{tretschk2018sequential} & SARL & off-policy &  & $ts$ & \change{\observation} &  & \whitec & \reachTarget{\stateRL} & \checkmark & \noAgent \\
         Mandlekar \etal \cite{mandlekar2017adversarially} & SARL & on-policy &  & $ts$ & \change{\stateRL} &  & \whitec & \minimize{\return} &  & \noAgent \\ 
         Zhang \etal \cite{zhang2020adaptive} & SARL & off-policy &  & $tr$ & \change{r} & \checkmark & \whitec & \learnTarget{\policy} &  & \noAgent \\    
         Huang \etal \cite{huang2019deceptive} & SARL & off-policy &  & $tr$  & \change{r} & \checkmark & \whitec / \blackc & \learnTarget{\policy} & \checkmark & \noAgent \\ 
         Liu \etal \cite{liu2021provably} & SARL & on/off-policy & & $tr$ & \change{\action} & \checkmark & \blackc & \learnTarget{\policy} &  & \surrogate \\ 
         Liu \etal \cite{liu2021provably} & SARL & on/off-policy & \checkmark & $tr$ & \change{\action} & \checkmark & \whitec & \learnTarget{\policy} &  & \noAgent \\ 
         Rakhsha \etal \cite{rakhsha2021reward} & MARL & on-policy & /\checkmark & $tr$ & \change{\action} & \checkmark & \grayc / \blackc & \learnTarget{\policy} & \checkmark & \extended \\         
         Majadas \etal \cite{majadas2021disturbing} & SARL & on-policy &  & $tr$ & \change{r} & \checkmark & \blackc & 
        \alter{\policy} &  & \noAgent \\         
         Kiourti \etal \cite{kiourti2020trojdrl} & SARL & on-policy &  & $tr$ & \change{\observation,r} & \checkmark & \whitec & \learnBackdoored{\policy} & ~ & \noAgent \\      
         Ashcraft \etal \cite{ashcraft2021poisoning} & SARL & off-policy &  & $tr$ & \change{\environment,r} & \checkmark & \blackc & \learnBackdoored{\policy} &  & \noAgent \\       
         Foley \etal \cite{foley2022execute} & SARL & on-policy &  & $tr$ & \change{\observation} & \checkmark & \whitec & \learnTargetWhenState{\policy} &  & \noAgent \\ 
         Ma \etal\cite{ma2018data} & SARL & off-policy &  & $tr$ & \change{r} & \checkmark & \whitec & \learnTargetWhenState{\policy} &  & \noAgent \\  
         Xu \etal \cite{xu2021transferable} & SARL & off-policy &  & $tr$ & \change{\environment} & \checkmark & \whitec / \blackc & \learnTargetWhenState{\policy} & \checkmark & \extended \\ 
        Zhang \etal \cite{zhang2021robust} & SARL & on-policy &  & $ts$ & \change{\observation} &  & \blackc & \minimize{\return} & \checkmark & \extended \\ 
        Tanev \etal \cite{tanev2021adversarial} & SARL & on/off-policy &  & $ts$ & \change{\environment} &  & \blackc & \minimize{\return} &  & \noAgent \\ 
        
        Lee \etal \cite{lee2020spatiotemporally} & SARL & on/off-policy &  & $ts$ & \change{\action} &  & \whitec & \minimize{\return} &  & \noAgent \\ 
        Lee \etal \cite{lee2020spatiotemporally} & SARL & on/off-policy &  & $ts$ & \change{\action} & ~ & \whitec & \minimize{\return} & \checkmark & \extended \\ 
        Huang \etal \cite{huang2017adversarial} & SARL & on/off-policy &  & $ts$ & \change{\environment} &  & \whitec / \blackc & \minimize{\return} &  & \noAgent \\         
        Buddareddygari \etal \cite{icra-22} & SARL & on-policy & & $ts$ & \change{\environment} &  &  \whitec & \reachTarget{\stateRL}  &  & \noAgent \\ 


        Wang \etal \cite{wang2021backdoorl} & MARL & on-policy &  & $tr$ & \change{\environment} & \checkmark & \whitec & \learnBackdoored{\policy} & \checkmark & \within \\ 
        Yu \etal \cite{yu_dont_2022} & MARL & off-policy &  & $tr$ & \change{\reward} &  \checkmark & \whitec & \learnBackdoored{\policy} & \checkmark   & \within \\
        Behzadan \etal \cite{behzadan2017vulnerability} & SARL & off-policy &  & $tr$ & \change{\environment} & \checkmark & \grayc & \learnTarget{\policy} &  & \surrogate \\ 
        
        Lin \etal \cite{lin2017tactics} & SARL & on-policy &  & $ts$ & \change{\observation} &  & \whitec & \minimize{\return} & & \extended \\ 
        Lin \etal \cite{lin2017tactics} & SARL & off-policy & \checkmark & $ts$ & \change{\observation} &  & \whitec & \reachTarget{\stateRL} & \checkmark & \extended \\ 
        Tekgul \etal \cite{tekgul22-esorics} & SARL & on/off-policy &  & $ts$ & \change{\environment} &  & \whitec & \minimize{\return} &  & \noAgent \\ 
        Russo \etal \cite{russo2019optimal} & SARL & off-policy &  & $ts$ & \change{\observation} &  & \blackc & \minimize{\return} & \checkmark & \extended \\ 
        Kos \etal \cite{kos2017delving} & MARL & on-policy &  & $ts$ & \change{\observation} &  & \whitec & \minimize{\return} &  & \noAgent \\ 
        Sun \etal \cite{sun2021strongest} & SARL & on/off-policy &  & $ts$ & \change{\observation} &  & \whitec & \minimize{\return} &  & \extended \\ 
        Korkmaz \etal \cite{korkmaz_nesterov_2020} & SARL & off-policy &  & $ts$ & \change{\stateRL} &  & \whitec & \minimize{\return} &  & \noAgent \\ 
        Inkawhich \etal \cite{inkawhich2020snooping} & SARL & on/off-policy &  & $ts$ & \change{\observation} &  & \blackc & \minimize{\return} &  & \surrogate \\ 
        Chen \etal \cite{chen2021stealing} & SARL & on/off-policy &  & $ts$ & \monitor{\action} &  & \blackc & \steal{\model} & \checkmark & \noAgent \\ 
        Yoon \etal~\cite{yoon2025real} & SARL & on-policy &  & $ts$ & \change{\stateRL} & & \grayc & \reachTarget{\stateRL} & \checkmark & \within\\
        Fan \etal~\cite{fan2025less} & SARL & on-policy & & $ts$ & \change{\observation} & & \blackc & \reachTarget{\stateRL} & \checkmark & \within\\
        
        Li \etal~\cite{li25-nn} & MARL & on/off-policy  &  & $ts$& \change{\observation} & &  \blackc & \minimize{\return} & & \within \\ 
        Bai \etal~\cite{Bai25-aaai} & SARL & off-policy/offline & & $ts$ & \change{\stateRL} &&  \grayc & \minimize{\return} & &  \extended \\
        Zheng \etal~\cite{Zheng24-dsn} & SARL / MARL & on-policy & & $ts$ & \change{\observation} && \blackc & \minimize{\return} & & \extended / \within \\ 
        Zhou \etal~\cite{zhou24-tsmc} & MARL & off-policy & & $ts$ & \change{\observation} & & \whitec / \blackc & \minimize{\return} & & \extended \\ 

        Ergu \etal~\cite{ergu25-TVT} & MARL & on-policy & & $ts$ & \change{\stateRL} & & \whitec & \minimize{\return} & & \extended \\ 
        Qu \etal~\cite{Qu2023AdvAttacks} & SARL & on-policy & & $ts$ & \change{\observation} & & \blackc & \reachTarget{\stateRL} & \checkmark & \within \\
        Alzubaidi \etal ~\cite{Alzubaidi2025multiagentcooperative} & MARL & on-policy &  & $ts$ & \change{\environment} & & \blackc & \reachTarget{\stateRL} & \checkmark & \within \\ 
        Alzubaidi \etal ~\cite{Alzubaidi2025multiagentcooperative} & MARL & on-policy &  & $ts$ & \change{\environment} & & \blackc & \minimize{\return} & \checkmark & \within \\ 

        \bottomrule
    \end{tabular}
    }
\end{table*}

\subsection{Test Time Attacks} The attacks that manipulate only test data cause either integrity or privacy violations. 

\myparagraph{Integrity Violations.} These attacks refer to an adversarial intervention at deployment (inference) time that causes the agent to make incorrect or unsafe decisions, violating the intended integrity of its policy, without altering its training process. Integrity violations include:
\begin{itemize}
    \item Target-State Manipulation Attack: The victim agent in the environment is perturbed by the attacker to reach a certain malicious target state: \reachTarget{\stateRL}. In other words, steering the agent to a specific attacker-chosen target state, \eg, driving to an unintended location such as a dead end, an enemy base, etc. The agent may still think it's acting optimally, but is being misled.
    \item Reward Minimization Attack: The victim agent is perturbed to perform an action that minimizes total return: \minimize{\return}, which potentially leads to unsafe and erratic behavior; \eg, veering the vehicle off the road or crashing into the guardrails. These attacks aim to make the agent fail its own reward function.
\end{itemize}
To perform these attacks, the attacker should have the ability to change (c) some \emph{test} inputs. 
Most of the attacks are Reward Minimization Attacks that alter a component of the system of an ego agent: state, action, environment, or observations, to minimize expected reward~\cite{zhao2020blackbox, xiang2021backdoor,russo2019optimal,huang2017adversarial,tanev2021adversarial,mandlekar2017adversarially,lee2020spatiotemporally,yang2020enhanced,zhang2021robust,lin2017tactics,inkawhich2020snooping,li25-nn,Bai25-aaai,Zheng24-dsn, zhou24-tsmc}. 
The attacker often modifies a system component only slightly, applying a perturbation bounded with an $l_p-$ norm constraint \cite{lin2017detecting, tanev2021adversarial, zhang2021robust}, making the attack less detectable. However, in~\cite{russo2019optimal}, the authors devise attacks by formulating the problem as MDP from the attacker's perspective. The attack is trained using the Deep Deterministic Policy Gradient (DDPG) with a reward function opposite of the agent to optimally undermine the agent's reward at test time.
The authors of \cite{tanev2021adversarial} exploit adversarial patches \cite{brown17-arxiv}, that is, small stickers containing evident perturbations. 
The advantage of adversarial patches is that they can be printed and used to alter the environment; therefore, attackers can perpetrate this attack even if they do not have access to the sensors. 
The authors of \cite{tanev2021adversarial} consider a system trained to grasp objects based on visual input, showing that the agent cannot perform its task in the presence of the patch. 

Few works alter a component of the system to cause the environment to induce the Target-State Manipulation Attack \cite{xiao2019characterizing, boloor2019simple, icra-22, sun2020stealthy, tretschk2018sequential, lin2017tactics, yoon2025real, fan2025less}. For example, in~\cite{tretschk2018sequential}, the attacker misguides the agent towards a different positive reward, rather than simply reducing its performance or making it fail. One of the attacks proposed in~\cite{lin2017tactics}: Enchanting Attack, where the attacker focuses on luring the DRL agent to a specific, pre-determined target state using the generative model and planning algorithm that generates a sequence of actions to guide the agent towards the target.

In \cite{tekgul22-esorics}, the authors claim that test-time attacks against DRL, to be practical, should work without considering the current state of the system. 
The reason is that pre-computing an adversarial example for each possible state is not feasible in practice, as the possible states are often too many, and the attacker would not be able to compute the attack for the current state before the state is already changed. 
To solve this problem, they consider universal adversarial examples~\cite{moosavi17-cvpr}, which can be computed offline and are effective for different states. 
Only the work in~\cite{lin2017tactics} proposes test-time attacks that perturb the environment to reach the attacker's desired state.

In multi-agent systems, the number of attack algorithms is quite limited. Even though some SARL attacks can be extended to MARL settings, where each agent is attacked individually, they do not necessarily account for interactions among agents.
The existing MARL attacks either manipulate some system components to minimize the expected reward~\cite{gleave2019adversarial,kos2017delving,li25-nn,Zheng24-dsn,zhou24-tsmc} or induce a backdoor~\cite{wang2021backdoorl}.
The authors of the work in~\cite{gleave2019adversarial} consider a two-player competitive DRL system and exploit the behavior of an agent within the environment (the opponent) to make the ego agent unable to win the match using an adversarial policy. One of the two attacks in~\cite{sun2020stealthy} is Antagonist Attack, which is domain-agnostic and general enough to be applied to a multi-agent environment; however, the evaluation is conducted in a single-agent environment. The attack proposed in~\cite{Zheng24-dsn} addresses both SARL and MARL. The attacker directly injects perturbations into the victim policy's inputs, whereas in multi-agent settings, the adversary controls an opponent agent to indirectly influence the victim's observations. In~\cite{li25-nn}, the attacker introduces a single adversarial agent into the cooperative MARL environment. This adversarial agent learns a policy that allows it to execute physically plausible actions that directly influence the observations of the victim agents. The attack is designed to be``unilateral'', \ie, the adversary influences the victims without being influenced by them in return. 

\myparagraph{Privacy Violations.} The attacks that cause privacy violations proposed so far monitor (m) the victim's behavior to clone its policy. 
This attack is called \emph{model stealing} ($\steal{\set M}$)~\cite{orekondy19-cvpr}. 
The only work in this direction is~\cite{chen2021stealing}, where the authors proposed an attack that aims to extract a proprietary DRL model, meaning to recover it with high fidelity and accuracy, based solely on observing its actions in an environment with black-box access.
In particular, the authors first identify the training algorithm family and then perform model extraction using imitation learning.
Imitation learning is a well-established approach for learning sequential decision-making policies. 
The desiderata for the copied model can be (1) accuracy, namely the ability to match or exceed the accuracy of the original model; (2) fidelity, namely a similar behavior to the original model (even committing the same errors). The authors showed that the proposed attack achieves high accuracy and fidelity.

The literature on the privacy of the DRL model is scarce and calls for more in-depth analysis.

\begin{table*}[t]
\caption{Attacks evaluated on autonomous driving tasks. 
/\checkmark indicates both the presence and absence of the feature in that paper. CDS stands for cooperative driving scenario.
}
\label{tab:ADAttacks}
\resizebox{0.99\textwidth}{!}{%
\begin{tabular}{r|rrrrrrr}
\toprule
\textbf{References} & \textbf{Target} & \textbf{Attacker's} & \textbf{Attacker's} & \textbf{Poisoning} & \textbf{Sequential} & \textbf{Multi} & \textbf{CDS} \\
& \textbf{Component} & \textbf{Goal} & \textbf{Action} & & \textbf{Attack} & \textbf{Agent} &\\
 \midrule
Wang \etal \cite{wang2020stop} &  controller &  traffic jam/crash &  positions and speeds &  \checkmark & &  \checkmark & \\
Ma \etal \cite{ma2019policy} & planning & car is rerouted & training rewards & \checkmark &  & & \\
Yang \etal \cite{yang2020enhanced} & end-to-end & car leaves lane & sensor input / FGSM &  &   \checkmark & &\\
Boloor \etal \cite{boloor2019simple} & end-to-end & car leaves lane & paint pattern on street   &  & \checkmark & & \\
Sun \etal \cite{sun2020stealthy} & end-to-end & car leaves lane & sensor environment / FGSM &  &   \checkmark & \checkmark & \\
Xiao \etal \cite{xiao2019characterizing} & end-to-end & suboptimal behavior & paint pattern on street   &  & /\checkmark & &\\
Buddareddygari \etal \cite{icra-22} & end-to-end & car leaves lane & add object &  &   &  &\\
Yu \etal \cite{yu_dont_2022} & end-to-end & car collision & car behavior & \checkmark &  \checkmark & \checkmark &\\
Yoon \etal \cite{yoon2025real} & end-to-end & car is misguided & sensor input/online perturbations & & \checkmark & &\\

Fan \etal~\cite{fan2025less} & end-to-end & car collision & sensor input & & \checkmark & &\\

Ergu \etal~\cite{ergu25-TVT}  & communication & erroneous allocation of resources & changes the perceived state & & & \checkmark & \checkmark\\
Qu \etal~\cite{Qu2023AdvAttacks} & Traffic Control & minimize travel time for colluded vehicles & change the perceived observation & & \checkmark & & \checkmark \\
Alzubaidi \etal~\cite{Alzubaidi2025multiagentcooperative}
 & end-to-end & maximize collision occurrence & drive system towards unsafe states & & \checkmark & \checkmark & \checkmark \\ 
Alzubaidi \etal~\cite{Alzubaidi2025multiagentcooperative}
 & end-to-end & minimize average speed of the vehicles & degrade cumulative reward indirectly & & \checkmark & \checkmark & \checkmark \\

 \bottomrule
\end{tabular}
}
\end{table*}

\subsection{Training Time Attacks} To perform these attacks, the attacker must be able to manipulate \emph{training} data, including observations, environment, or rewards, either before or during learning. Existing training-time attacks target availability or integrity violations.

\myparagraph{Availability Violations.}
The goal of these attacks is to disrupt or degrade the victim's ability to learn the legitimate policy. This can be done by:
\begin{itemize}
    \item Target Manipulation Attack: Forcing the agent to learn an attacker-chosen policy: \emph{$\learnTarget{\policy}$}. The model may still converge, but to an unusable or bad policy. For example, the attacker may poison the environment or reward function so that actions taken for left-side driving are consistently rewarded instead of the legitimate right-side driving. Here, the agent learns a consistent policy but for the wrong traffic system.
    \item Untargeted Policy Attack: Preventing the agent from learning the intended policy without enforcing a specific alternative: \emph{$\alter{\policy}$}. This can be done by altering the reward or overall return.
\end{itemize}

In \cite{majadas2021disturbing}, the authors propose a black-box un-targeted policy attack. They show how disturbances in the reward function affect the convergence of a learning agent.
They assume that the attacker does not have any knowledge about the learned policy, but can occasionally flip the sign of the reward when the environment reaches a chosen state. Their work focuses on on-policy DRL and analyzes different exploration strategies.
Their experimental analysis shows that small exploration probabilities are more resilient to perturbations of the reward. 
In \cite{ma2019policy, zhang2020adaptive, huang2019deceptive, rakhsha2020policy}, the authors alter the reward to force the agent to learn a target policy chosen by the attacker, \eg, to make a robot learn to reach a location chosen by the attacker instead of the location desired by the developer of the RL system.
In \cite{behzadan2017vulnerability}, the authors propose an ad-hoc manipulation of the environment at training time by an adversarial agent. The attacker first trains its own adversarial DQN using the desired adversarial policy $\policy_{adv}$, then uses it to generate perturbed samples that attack the environment of the target system. In a multi-agent setting, only~\cite{rakhsha2021reward} is proposed, where the authors focus on \emph{population learner} black-box scenario. While each individual learner may be a single agent, the overall attack framework is designed for a multi-agent context in which the attacker seeks to manipulate a series of independent learners. The attack itself, though, is model-agnostic, however, the proposed method unfolds in two phases: (i) an exploration phase, which gathers the information from the environment to estimate a set of plausible \model; and an (ii)  attack phase, in which the attacker uses the estimated model of the environment to determine the optimal reward perturbations, therefore, marked ``/\checkmark" for this paper.

\myparagraph{Integrity Violations.} The attacks that cause an integrity violation apply the following tactics: 
\begin{itemize}
    \item Backdoor Poisoning Attack: Inserting a trigger pattern into training data \emph{$\learnBackdoored{\policy}$} so the agent behaves normally except when the trigger appears. For example, an autonomous car may drive correctly but act unpredictably when a stop sign has a yellow sticker.
    \item Targeted Poisoning Attack: Manipulating training so the agent learns a malicious policy only in specific states \emph{$\learnTargetWhenState{\policy}$}, such as roundabouts or intersections, while behaving normally elsewhere. For instance, the attacker may poison training so the agent consistently exits a roundabout incorrectly, while in other states, the agent has a good driving policy.
\end{itemize}
Training-time integrity attacks are powerful because they typically preserve normal behavior, making them difficult to detect through observation. In targeted poisoning attacks, abnormal behavior occurs only under specific triggers or states. For example, in~\cite{xu2021transferable}, carefully designed training-time perturbations cause the agent to take a desired action in certain states while behaving normally otherwise. Similarly, in~\cite{ma2018data}, the paper target contextual bandits, a class of RL algorithms suited for dynamic environments with rapidly changing, limited action sets.
Several works study backdoor poisoning. In~\cite{kiourti2020trojdrl}, training samples with a trigger assign high rewards to certain actions, causing random behavior when triggered. Similarly, in~\cite{ashcraft2021poisoning}, the agent learns a target policy under trigger presence. They also introduce in-distribution triggers, patterns that can occur due to agent-environment interactions that are hard to detect, and show success with only 10–20\% poisoned samples.
In multi-agent settings, only backdoor attacks have been proposed. In BackdoorRL~\cite{wang2021backdoorl}, the victim agent is trained using imitation learning from a mix of ``normal" and ``backdoored" trajectories. The backdoor is then activated by a series of specific trigger actions performed by the adversary agent.
In~\cite{yu_dont_2022}, the backdoor is triggered by an agent near the victim. 
\cite{wang2020stop} considers a traffic congestion control system, and assumes that the attacker uses the state of the car agents in the neighborhood of the ego agent to trigger the backdoor. 

\begin{figure*}[t]
    \centering
    \includegraphics[width=0.85\textwidth]{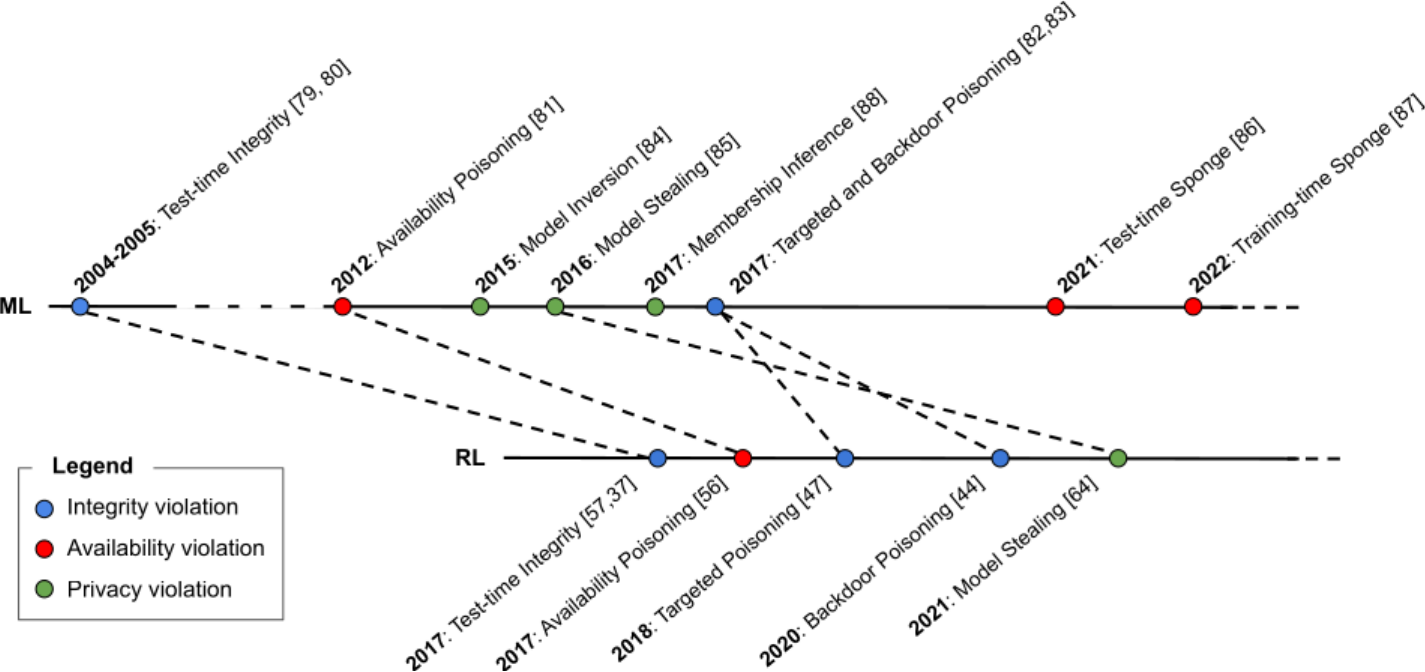}
    \caption{Timeline of attacks against reinforcement learning, compared to those proposed against machine-learning classifiers.}
    \label{fig:timeline}
\end{figure*}

\subsection{Metrics to Measure the Effectiveness of the Attack}\label{sec:attack_metrics}
The evaluation of attacks can be conducted through metrics that assess the generic attack performance, such as Attack Success Rate (ASR), cumulative reward degradation, and policy performance drops.
Many works also report perturbation magnitude constraints and transferability across models, mirroring established adversarial evaluation protocols.
Other metrics can instead be specific to the AD scenario and are incorporated into the simulation platforms.
For example, the collision rate, time-to-collision, crash rate, or number of traffic violations assess the impact of the attack in the context of the specific application.
Importantly, realistic assessment should also consider real-time perturbation generation constraints, sensor noise and environmental variability, latency, physical-world feasibility of perturbations, and performance in multi-agent traffic scenarios. 
Such factors are largely absent from the existing work analyzed in this survey. 

\subsection{Attacks against Autonomous Driving}
\label{sec::ADAttacks}

This section reviews attacks on autonomous vehicles, highlights the properties needed for real-world threats, and discusses current limitations and future research directions.

\myparagraph{State of the Art.} 
Table~\ref{tab:ADAttacks} summarizes attacks on AD. 
Most attacks target end-to-end systems at test time, aiming to force lane departures or unsafe maneuvers~\cite{yang2020enhanced, boloor2019simple, sun2020stealthy, xiao2019characterizing}, sometimes via physical perturbations such as painted road lines or printed artifacts. Others exploit sensor inputs, environmental responses, or online image streams~\cite{yoon2025real}, and some leverage sparse perturbations to evade detection~\cite{fan2025less}. 
In end-to-end RL, perturbing the observation simultaneously corrupts perception and planning.
Training-time threats include policy poisoning, which manipulates rewards to misguide planning~\cite{ma2019policy}, and backdoor attacks that target either traffic optimization algorithms~\cite{wang2020stop} or vehicle behaviors~\cite{yu_dont_2022}. 
Importantly, these attacks exploit AD's unique characteristics. The environment is not closed and fully controllable as in, \eg, robotic control or simpler scenarios.
In most of the analyzed threats, adversarial actions are tied to specific interactions with the physical environment to affect the agent's partial observability. In this scenario, controlling and ensuring that the environment is not manipulated is much more challenging, as any malicious actor might gain access to the operational area, sensors, or infrastructure.
The outcome of an attack is also more difficult to assess: benign external agents, such as other vehicles, cyclists, or pedestrians crossing the road, may both influence and be affected by the compromised behavior.
Moreover, the real-time constraints of autonomous driving systems and the tight feedback loop between perception, planning, and control limit the opportunity for detection or recovery once an attack is triggered. Some recent papers proposed attacks in cooperative driving setups. In~\cite{Qu2023AdvAttacks}, colluded vehicles attacks a DRL based adaptive traffic control system (ATCS) by duplicating their presence and report the falsified traffic conditions, in order to achieve minimum total travel time. In~\cite{ergu25-TVT} the authors propose an attack against the communication system employed by the CDS. \cite{Alzubaidi2025multiagentcooperative} proposed 2 attacks: a collision attack and a speed reduction attacks that hampers the cooperative goals of CDS.

Collectively, these studies reveal diverse attack surfaces in AD but remain largely simulator-based, highlighting the urgent need for real-world evaluations to assess their practical feasibility and safety implications.

\myparagraph{Practical Shortcomings.} 
While current RL attacks serve as proofs of concept, their impact on real autonomous vehicles remains uncertain. Most are not tested on actual cars, likely due to the high cost of AVs, and only a few consider actions feasible in the physical world~\cite{boloor2019simple,xiao2019characterizing}. 
Real-world operational constraints, including sensor noise, calibration errors, latency, and environmental variability such as weather, lighting, and road surface conditions, are often deemed outside of the scope of these studies. 
However, these factors can significantly affect both the feasibility and efficacy of adversarial manipulation, particularly in safety-critical, real-time driving scenarios.
Critical practical aspects, such as computation time for real-time perturbations, attack transferability across models, and the feasibility of sequential versus single-shot attacks, are largely unexplored. Finally, only a minority of attacks operate in realistic black-box settings, limiting the assessment of their applicability to proprietary AV models. These gaps highlight the need to evaluate RL attacks under real-world constraints to understand their true risk in AD scenarios.



\subsection{Adversarial RL Timeline}

Attacks against RL are adaptations of attacks previously proposed against standard ML systems, particularly against classification tasks. In Figure \ref{fig:timeline}, we highlight the connections between the attacks proposed against standard ML systems and those proposed against RL. 
The first attacks against RL were test-time integrity attacks \cite{lin2017tactics, mandlekar2017adversarially}, proposed in 2017, which could cause the system to misbehave when a carefully crafted sample was provided as input. These attacks are inspired by those \cite{dalvi04, lowd05}, which were proposed more than ten years earlier, between 2004 and 2005, against ML classifiers that misclassified carefully crafted adversarial samples. For example, misclassifying a spam email as legitimate. The first training-time poisoning attack against RL was also proposed in 2017 \cite{behzadan2017vulnerability}. The goal of this attack was to make the system unable to work correctly, specifically by manipulating the target policy, causing a denial of service. In this case, the attack is also inspired by an attack \cite{biggio12-icml} proposed earlier (2012) against ML classifiers and, in particular, against the Support Vector Machine, affecting its ability to learn to correctly classify the samples. In 2018 and 2020, researchers proposed the first integrity-violation train-time attacks against RL. In particular, the first paper on targeted poisoning attack~\cite{ma2018data} in 2018 and backdoor poisoning attack~\cite{kiourti2020trojdrl} in 2020 were proposed.
Both were inspired by articles proposed in 2017 \cite{koh17-icml, gu17}, the former aimed to misclassify test samples by manipulating a few training samples, while the latter made the classifier learn to misclassify all samples containing a specific trigger. The only attack that violates the privacy of RL systems performing model stealing~\cite{fredrikson15-ccs}, was proposed recently in 2021 and is inspired by an attack previously proposed against ML classifiers \cite{tramer16-usenix} in 2016. An attack against ML to determine whether some data were used to train the model \cite{fredrikson15-ccs} and an attack to obtain a copy of the model \cite{tramer16-usenix} were proposed, respectively, in 2015 and 2016. Attacks aimed at slowing down ML systems have been proposed \cite{shumailov21-eurosp,cina_energy-latency_2022}, but have never been demonstrated against RL systems.

\subsection{Future Research Directions}
While many attacks have been explored against ML classifiers, their applicability to RL remains underexplored. Future research should address security violations in RL by tailoring threat models to application-specific needs. Beyond model theft, privacy attacks such as model inversion \cite{fredrikson15-ccs} and membership inference \cite{shokri_membership_2017} could expose sensitive driving data. Availability attacks, such as sponge attacks that delay system responses, are especially critical in real-time AD.
As a safety-critical domain, AD demands realistic attack evaluations that go beyond simulations. Black-box attacks and transferability from simulated to real vehicles need to be studied. 
As AD systems increasingly shift toward end-to-end learning, new attack surfaces emerge across the training and decision-making pipelines. 
In contrast to modular systems, where vulnerabilities can be analyzed within relatively isolated components, end-to-end models propagate perturbations through shared representations and closed-loop decision-making, potentially amplifying small input manipulations into unsafe control actions. 
The end-to-end integration increases the difficulty of tracking the source of errors, as they might be introduced at the perception level and only manifest at the control stage. 
Securing such pipelines, therefore, requires dedicated strategies that combine robust representation learning, safety-constrained policy optimization, runtime anomaly detection, and fail-safe control mechanisms. 
Future research should investigate cross-module error propagation, resilience to distribution shift, and real-time safety guarantees.

\section{Defenses}
\label{sec:defenses}
Various defenses have been proposed to make the DRL more robust against attacks. In the following, we first explain how state-of-the-art DRL defenses can be categorized. 
We then focus on defenses tailored to the autonomous-driving environment, even though only a single such defense has been proposed. Thus, we discuss open challenges and conclude the section with a review of potential future directions for the development of defenses. 

All existing DRL defense methods have adopted previously developed techniques for ML classification tasks.
They can be divided into three main categories: a) those that try to counteract existing attacks, \ie \emph{reactive} defenses, and b) those that act to prevent future attacks, \ie \emph{proactive} defenses~\cite{biggio18}, and c) those that detect the attack and provide proactive solutions, \emph{hybrid}. Most defense techniques can be categorized as one or more of the following:


\begin{itemize}
    \item Adversarial training (\advtraining, proactive): Iteratively re-train the system on simulated attacks. These defenses are heuristic and do not have formal guarantees. 
    \item Game Theory (\gametheory, proactive): Model the interaction between the agent and adversary as a game (Nash equilibrium~\cite{bruckner12} or Stackelberg game~\cite{liu10a}) and solve the robust policy, \eg using minimax optimization. 
    They are more principled than adversarial training; however, they are computationally costly. 
     \item Regularization (\regularization, proactive): Adds penalty terms during training to penalize the agent's sensitivity to input perturbations or choosing the best action considering a possible worst-case perturbation of the input. 
    \item Distillation (\distillation, proactive): This technique was originally proposed for model compression~\cite{hinton2015distilling}. Transfer knowledge from a robust, larger network, the teacher model, to a smaller or less complex, the student model, often smoothing decision boundaries. 
    \item Ensemble (\ensemble, proactive): Use multiple agents and aggregate their outputs to make decisions more robust against perturbations or adversarial attacks. For example, an ensemble of DQN agents, each trained on different random seeds and perturbation strategies. The final action can be based on consensus.
    \item Detection (\detection, reactive): The goal is to identify whether the input to the RL agent (\eg, observations, environment dynamics, etc.) has been adversarially perturbed or is out-of-distribution. These test samples are then flagged for further processing.
    \item Sanitization (\sanitization, reactive): It is a test-time defense that pre-processes the input to remove adversarial perturbations before feeding them into the RL system.
\end{itemize}

\subsection{Categorizing the Defenses against RL}

In this paper, we aim to provide a taxonomy that is useful to both researchers and practitioners. 
Thus, we consider not only the peculiarities of the defenses, but also the ones of the RL systems on which they have been shown to be effective. 
Each proposed defense can be categorized into the defense model and the considered threat model:

\myparagraph{Defense Model.} Characteristics of the defense technique.
\begin{itemize}
\item \emph{Single vs. Multi Agent}: Based on the number of agents considered in the environment, the proposed methods can be evaluated for either SARL or MARL environments.
\item \emph{Policy Update.} The frequency with which the policy is updated: on-policy, off-policy, off-line.
\item \emph{Defense Type} Based on the timeline where the defense is deployed, it can be proactive, reactive, or hybrid.
\item \emph{Defense Technique.} The specific technique used in the paper, \ie, adversarial training, detection, etc.
\end{itemize}
\myparagraph{Threat Model.} Characteristics of the attack model that the defender used to defend their agent. Same as in sect.~\ref{sec: attack_category}.


\begin{table*}[!ht]
    \caption{Defenses for reinforcement learning systems. The presence of the \checkmark indicates that the corresponding attribute is true for the defense. For the attacker’s knowledge, we use \blackc \ to represent black-box,  \grayc \ for gray-box, and \whitec \ for white-box. All the defenses proposed so far have been tested on model-free RL algorithms, except~\cite{Bhardwaj23-NEURIPS}.}\label{tab:defenses}
    \centering
    \resizebox{0.99\textwidth}{!}{%
    \begin{tabular}{r|rrrrrrrrrrr}
    \toprule
        \toprule
       & \multicolumn{3}{c}{\textbf{DEFENSE MODEL}} & \multicolumn{7}{c}{\textbf{THREAT MODEL}}
        \\
        \cmidrule(l){2-5} \cmidrule(l){6-12}

        References &
       
       Single vs. &
       Policy &
       Defense &
       Defense &
       Attack's &
       Attacker's &
       Poisoning &
       Attacker's &
       Attacker's &
       Sequential &
       Attacker's \\
        & 
        
        Multi-agent &
        Update &
        Type &
        Technique &
        Time &
        Action &
        &
       Knowledge &
       Goal &
       Attack &
       Agent \\
        
        \midrule
        Everitt \etal \cite{everitt2017reinforcement} & SARL & on-policy & \emph{proactive} & \regularization & $tr$ & \change{r} & \checkmark & \grayc & \alter{\policy} &  & \noAgent \\ 
        Everitt \etal \cite{everitt2017reinforcement} & SARL & on-policy & \emph{reactive} & \sanitization & $tr$ & \change{r} & \checkmark & \grayc & \alter{\policy} &  & \noAgent \\ 
        Wang \etal \cite{wang2021backdoorl} & SARL & on-policy & \emph{reactive} &  \sanitization & $tr$ & \change{\environment} & \checkmark & \whitec & \learnBackdoored{\policy} & \checkmark & \within \\ 
        Wang \etal \cite{wang2020reinforcement} & SARL & off-policy & \emph{reactive} & \sanitization & $tr$ & \change{r} & \checkmark & \blackc & \alter{\policy} &  & \noAgent \\ 
        Lin \etal \cite{lin2017detecting} & SARL & on/off-policy & \emph{hybrid} & \detection & $ts$ & \change{\observation} &  & \whitec & \minimize{\return} & \checkmark & \noAgent \\ 
        Han \etal \cite{han2020adversarial} & SARL & on/off-policy & \emph{reactive} & \detection + \sanitization & $tr$ &  \change{\stateRL,r} & \checkmark & \blackc & \learnTargetWhenState{\policy} & \checkmark & \surrogate \\ 
        Havens \etal \cite{havens2018online} & SARL & on-policy & \emph{reactive} &  \detection + \sanitization & $tr$ &  \change{\observation} &  & \blackc & \learnTarget{\policy} & \checkmark & \noAgent \\
        Garcıa \etal \cite{garcia2022instance} & SARL & on-policy & \emph{reactive} &  \detection + \sanitization & $ts$ & \change{\observation} &  & \whitec & \minimize{\return} &  & \noAgent \\ 
        Pinto \etal \cite{pinto2017robust} & SARL & on-policy & \emph{proactive} & \gametheory & $tr$ & \change{\environment} &  & \whitec & \alter{\policy} & \checkmark & \extended \\  
        Tessler \etal \cite{tessler2019action} & SARL & on/off-policy & \emph{proactive} & \gametheory & $tr$ &  \change{\action} &  & \whitec & \alter{\policy} & \checkmark & \extended \\ 
        Banihashem \etal \cite{banihashem2021defense} & SARL & on-policy & \emph{proactive} & \advtraining & $tr$ & \change{\reward} & \checkmark & \whitec & \learnTarget{\policy} & \checkmark & \noAgent \\ 
        Tan \etal \cite{tan2020robustifying} & SARL & on-policy & \emph{proactive} &  \advtraining & $tr$ & \change{\action} &  & \whitec & \learnTarget{\policy} &  & \noAgent \\ 
        Mandlekar \etal \cite{mandlekar2017adversarially} & SARL & on-policy & \emph{proactive} &  \advtraining & $ts$ &  \change{\environment,\observation} &  & \whitec & \minimize{\return} &  & \noAgent \\ 
        Pattanaik \etal \cite{pattanaik2017robust} & SARL & off-policy & \emph{proactive} & \advtraining  & $ts$ & \change{\observation} &  & \whitec & \reachTarget {\stateRL } &  & \noAgent \\ 
        Lee \etal \cite{lee2021query} & SARL & on-policy & \emph{proactive} & \advtraining & $ts$ & \change{\action} &  & \blackc & \reachTarget{\stateRL} & \checkmark & \extended \\ 
        Rajeshwaran \etal \cite{RajeswaranGRL17} & SARL & on-policy & \emph{hybrid} & \ensemble + \advtraining & $tr$ & \change{\environment} &  & N.A & N.A & N.A & N.A. \\ 
        Zhang \etal \cite{zhang2021robust} & SARL & on-policy & \emph{proactive} &  \advtraining & $ts$ & \change{\observation} &  & \blackc & \minimize{\return} & \checkmark & \extended \\ 
        Zhang \etal \cite{zhang2020robust} & SARL & on/off-policy & \emph{proactive} & \regularization & $ts$ & \change{\environment} &  & \whitec & \minimize{\return} &  & \extended \\ 
        Wu \etal \cite{wu2022crop} & SARL & off-policy & \emph{proactive} & \regularization & $ts$ & \change{\observation} &  & \whitec & \minimize{\return} & \checkmark & \noAgent \\ 
        Lutjens \etal \cite{lutjens2020certified} & SARL & off-policy & \emph{proactive} & \regularization & $ts$ & \change{\observation} &  & \whitec & \reachTarget {\stateRL } &  & \noAgent \\ 
        Oikarinen \etal \cite{oikarinen2020robust} & SARL & on/off-policy & \emph{proactive} & \regularization + \advtraining & $ts$ & \change{\observation} &  & \whitec & \minimize{\return} & \checkmark & \noAgent \\ 
        Qu \etal \cite{qu2020adversary} & SARL & off-policy & \emph{proactive} & \distillation & $ts$ & \change{\observation} &  & \whitec & \minimize{\return} &  & \noAgent \\ 
        Fischer \etal \cite{fischer2019online} & SARL & off-policy & \emph{proactive} &  \distillation + \advtraining & $tr$, $ts$ & \change{\observation} &  & \whitec & \alter{\policy}, \minimize{\return} & & \noAgent \\ 
        Kos \etal \cite{kos2017delving} & SARL & on-policy & \emph{proactive} &  \advtraining & $ts$ & \change{\observation} &  & \whitec & 
        \minimize{\return} & \checkmark & \noAgent \\ 
        He \etal \cite{he2024trustworthy} & SARL & off-policy & \emph{proactive} & \advtraining  & $ts$ & \change{\observation} &  & \whitec & \minimize{\return} & \checkmark & \noAgent \\
        Wang \etal~\cite{wang2024explainable} & SARL & on-policy & \emph{hybrid} & \detection + \advtraining & $ts$ & \change{\observation} &  & \whitec & \minimize{\return} & \checkmark & \noAgent \\
        Majadas \etal~\cite{majadas2024clustering} & SARL & off-policy & \emph{reactive} & \detection & $ts$ & \change{\observation} & & \whitec & \minimize{\return} &\checkmark & \extended \\
        
        Wang \etal~\cite{wang25-usenix} & SARL & off-policy & \emph{proactive} & \regularization & $ts$ & \change{\observation} & & \whitec/\blackc & \minimize{\return} & & \noAgent \\ 
        Liu \etal~\cite{Liu24-ijcnn} & SARL & off-policy & \emph{proactive} & \advtraining  & $ts$ & \change{\action} & & \whitec  & \minimize{\reward} & & \extended \\ 
        Bhardwaj \etal~\cite{Bhardwaj23-NEURIPS} & SARL & offline & \emph{proactive} & \advtraining  & $ts$  & \change{\policy} & & \whitec & \minimize{\return} & & \noAgent \\ 
        Korkmaz \etal~\cite{Korkmaz23-icml} & SARL & off-policy & \emph{reactive} & \detection & $ts$ & \change{\observation} & &  \whitec &  \minimize{\return} & & \noAgent \\ 
        Meng \etal~\cite{meng23-is} &   SARL & on-policy & \emph{proactive} & \advtraining  & $ts$ & \change{\observation} &  & \whitec & \minimize{\return} & & \noAgent \\ 
        Guo \etal~\cite{Guo25-TSMC} & MARL  & off-policy  & \emph{proactive} & \advtraining  & $ts$ & \change{\observation} & & \whitec & \minimize{\return} & & \within \\ 
        Wang \etal~\cite{wang2025MultiAgentRobust} & MARL & off-policy & \emph{proactive} & \regularization & $ts$ & \change{\observation} & & \whitec & \minimize{\return} & \checkmark & \within \\
        Bukharin \etal~\cite{bukharin23-neurips} & MARL & on/off-policy & \emph{proactive} & \gametheory  & $ts$ & \change{\observation} / \change{\action} &  & \whitec & \minimize{\return} & & \noAgent \\ 
        Zhou \etal~\cite{Zhou_2024_rofmac} & MARL & on-policy & \emph{proactive} & \regularization + \advtraining & $ts$ & \change{\observation} & & \whitec & \minimize{\return} & & \noAgent \\
        Wang \etal~\cite{Wang2025MARLDefense} & MARL & on-policy & \emph{proactive} & \regularization + \advtraining & $ts$ & \change{\observation} & & \whitec & \minimize{\return} & \checkmark & \noAgent \\

        \bottomrule
    \bottomrule
    \end{tabular}
    }
\end{table*}

\subsection{Defenses for RL Agents}

This section provides an overview of defenses for RL agents. Table \ref{tab:defenses} is a compact categorization of state-of-the-art defenses against single agents. Since all the defenses are model-free, we omit the model-based column. 
In the following, we describe the defenses in more detail, subdividing them into proactive, reactive, and hybrid types.

\myparagraph{Proactive.} Proactive defenses are applied during training to build inherent robustness into RL agents, preparing them to withstand adversarial conditions at test time. Techniques include adversarial training, game-theoretic methods, regularization, and distillation. \emph{Adversarial training} is widely used~\cite{banihashem2021defense, tan2020robustifying, mandlekar2017adversarially, pattanaik2017robust, zhang2021robust, kos2017delving, he2024trustworthy, Guo25-TSMC, Liu24-ijcnn, Bhardwaj23-NEURIPS, meng23-is}. Even though the mentioned defenses use adversarial training as an underlying technique, they defend against different threat models (see Table~\ref{tab:defenses}). For instance, Banihashem et al.\cite{banihashem2021defense} defend against reward poisoning attacks, while Tan et al.\cite{tan2020robustifying} focus on action-space and actuator perturbations. Zhang et al.\cite{zhang2021robust} and Kos et al.\cite{kos2017delving} target test-time attacks, Bhardwaj et al.\cite{Bhardwaj23-NEURIPS} train offline to improve policies relative to a reference, and Guo et al.\cite{Guo25-TSMC} and Wang et al.~\cite{Wang2025MARLDefense} extend adversarial training to MARL with state-level attacks.
\emph{Regularization-based} defenses\cite{everitt2017reinforcement, zhang2020robust, wu2022crop, lutjens2020certified, wang25-usenix} add constraints during training to (a) prevent overfitting to training data or (b) reduce sensitivity to observation perturbations. These approaches are certified, providing guarantees on the maximum reward change under bounded input perturbations. For instance, Everitt et al. \cite{everitt2017reinforcement} proposed quantilisation, in which the agent selects states from a top-reward quantile rather than always maximizing reward, introducing randomness and conservatism against adversarial manipulation. Zhang et al.\cite{zhang2020robust} added a penalty term to the RL objective to encourage robustness to state perturbations. More recently, Wang et al.\cite{wang2025MultiAgentRobust} combined risk estimation with constraint optimization to make MARL policies resilient to worst-case observation perturbations.
\emph{Distillation-based} defenses are relatively unexplored. Qu et al.\cite{qu2020adversary} proposed Adversary Agnostic Policy Distillation (A2PD), where a robust teacher guides a student policy via a novel loss, improving adversarial robustness without exposure to adversarial examples. In contrast, Fischer et al. \cite{fischer2019online} combined policy distillation with adversarial training: a Q-network is trained conventionally, while a student network mimics it and is adversarially trained, enhancing robustness while preserving performance.
\emph{Game-theoretic} defenses\cite{pinto2017robust, tessler2019action, bukharin23-neurips} frame training as a minimax game against adversaries, making them a form of adversarial training. Pinto et al. \cite{pinto2017robust} modeled the setup as a zero-sum game in which a protagonist learns the task while an adversary applies disturbance forces, yielding a protagonist policy robust to worst-case disturbances. Unlike most works, Bukharin et al.~\cite{bukharin23-neurips} addressed MARL, demonstrating the importance of Lipschitz continuity for robustness and proposing a Stackelberg game approach in which the policy (leader) anticipates the attacker's (follower) actions via smoothed optimization, enabling robust decision-making. 
In~\cite{lee2021query}, authors proposed a proactive defense approach based on \emph{transfer-learning} in coordination with adversarial training. In this paper, the authors describe a variant of adversarial training in which the weights of a pre-trained nominal policy serve as a starting point for further fine-tuning. By leveraging the prior weights, they aim to induce robust behaviors into the nominal policy. 
Lately, researchers have adopted a combination of different proactive approaches~\cite{oikarinen2020robust, Zhou_2024_rofmac, fischer2019online} to devise a strong defense system. For example, in~\cite{Zhou_2024_rofmac}, while the method generates adversarial perturbations at training time, the extra penalty term for action loss ensures that trained actors perform well on both clean and adversarial states. However, in~\cite{fischer2019online} the architecture is divided into two networks: policy network (student) and Q-network, where Q network learns the optimal Q-values and guides the overall learning process and remains largely unchanged, whereas the student network is the primary target for adversarial training and learns to mimic the policy of the Q-network through a process called policy distillation. In this work, both train-time and test-time are considered for the evaluation of the proposed defense.
We highlight that none of the surveyed MARL defenses (\cite{Guo25-TSMC,wang2025MultiAgentRobust,bukharin23-neurips,Zhou_2024_rofmac}) specifically target low overhead or efficient scalability in large multi-agent security settings. 

\myparagraph{Reactive.} Reactive defenses operate during testing or deployment, aiming to detect, reject, or correct adversarial inputs \emph{after} an attack occurs. Techniques include detection, sanitization, and reward monitoring. \emph{Detection-based} defenses~\cite{lin2017detecting, majadas2024clustering, Korkmaz23-icml} focus on identifying adversarial samples. Lin et al.\cite{lin2017detecting} used an action-conditioned visual foresight module to predict future frames, detecting attacks by comparing its action distribution with the agent’s. Korkmaz et al.\cite{Korkmaz23-icml} distinguished adversarial samples via local curvature analysis of the policy’s cost function, noting benign inputs exhibit larger negative curvature. Majadas et al.~\cite{majadas2024clustering} applied clustering of benign transitions, flagging deviations in real time as adversarial.
Whereas the defenses proposed in~\cite{wang2021backdoorl, wang2020reinforcement, han2020adversarial, everitt2017reinforcement} fall in~\emph{sanitization} category in which the model filters or reconstructs noisy/adversarial inputs before passing them on to the agent. For example, Wang et al.\cite{wang2020reinforcement} denoise inputs by estimating a reward confusion matrix and recovering unbiased surrogate rewards. Wang et al.\cite{wang2021backdoorl} sanitize trained models to remove backdoor functionality, though only in single-agent settings despite targeting multi-agent attacks. Han et al.\cite{han2020adversarial} propose an inversion defense that cancels perturbations by finding a corrective signal $\delta'$ such that, when added to the already perturbed state $s' + \delta$, effectively cancels out the original perturbation $\delta$. Everitt et al.\cite{everitt2017reinforcement} introduce “decoupled RL,” leveraging external, trustworthy information to rectify the reward interpretation.
Other reactive approaches can be a combination of \emph{detection \& sanitization}, where instead of sanitizing all input samples, these approaches first identify the adversarial/out-of-distribution sample and then re-actively sanitize it for the model~\cite{havens2018online, garcia2022instance}. Havens \etal~\cite{havens2018online} proposed an online and model-agnostic defense approach called Meta-Learned Advantage Hierarchy (MLAH), which leverages an advantage function to detect if the adversarial attack is present or not and then decides on which policy to choose. In~\cite{garcia2022instance}, however, the model uses a memory of past, safe states to detect and correct adversarial inputs at test time. 

\myparagraph{Hybrid} These defenses combine proactive training-time mechanisms with reactive test-time components to ensure robustness and adaptability during deployment. Lin et al.\cite{lin2017detecting} detect adversarial examples via action distribution comparisons and proactively suggest corrective actions to maintain performance. Rajeswaran et al.\cite{RajeswaranGRL17} use ensembles with adversarial training to learn policies robust to model errors, coupled with a model adaptation loop for reactive adjustments, though tested only against random perturbations. Wang et al.~\cite{wang2024explainable} integrate robust adversarial training against worst-case attacks with saliency-based detection, providing insights into the DRL agent’s decision-making based on sensor inputs.
It is worth noting here that most defense techniques are designed in a single-agent setting. Although the defenses can be scaled to a multi-agent setting where each agent has its own individual defense method deployed, the defenses may not necessarily consider the other RLs into account.

\subsection{Metrics to Measure the Effectiveness of the Defenses}\label{sec:defense_metrics}

For defenses, evaluation criteria can depend on whether the approach is proactive or reactive. Proactive defenses, such as adversarial training or robust policy learning, are typically assessed through robustness under attack, and the trade-off between performance under attack and in normal conditions. 
In contrast, reactive defenses, including anomaly detection and runtime monitoring, should be evaluated based on detection accuracy (e.g., false positive and false negative rates), response latency, recovery effectiveness after attack onset, resilience against adaptive adversaries, and the impact of intervention mechanisms (e.g., safe-stop behaviors) on overall safety and traffic flow. 
We also observe that existing studies on MARL defenses tend to neglect scalability analysis. 
Future evaluation frameworks and benchmarks should therefore explicitly measure computational and communication costs associated with scaling these defenses to large autonomous fleets.

\subsection{Defenses for Autonomous Driving}\label{sec:addefenses}

To the best of our knowledge, very few defenses have been evaluated in the context of AD: one on a simplified AD scenario, a straight road \cite{wu2022crop}, and another on intersection passing for a multi-agent scenario~\cite{wang2025MultiAgentRobust}. However, no mitigation has been tested on a real vehicle.
One possible reason could be the cost of an actual vehicle to evaluate such defenses, which easily amounts to more than 20,000\$.\footnote{\url{https://qz.com/924212/what-it-really-costs-to-turn-a-car-into-a-self-driving-vehicle/}}.
This cost is not affordable for most research labs.
Albeit using a simulator can, to some degree, solve this problem. There are other challenges that defenses should face to be applicable in AD, such as:
\begin{itemize}
    \item Requirement of only hardware that is usually already on board~\cite{shen2022sok}. For example, it is unlikely that an autonomous vehicle based solely on cameras will be equipped with a LiDAR sensor only for defense purposes. 
    \item Evaluation considering all components of the AD system~\cite{shen2022sok}. There is a possibility that the defense interacts with other parts of the car, which has to be taken into account during development.
\end{itemize}
In contrast, some requirements can also be approximated in a simulator without involving an actual car. These include the introduced overhead, which cannot be too large as otherwise the car cannot react properly anymore. Some of the proactive defenses discussed in this section (for example, when based on regularization~\cite{lin2017detecting, wang2020reinforcement, everitt2017reinforcement} or distillation~\cite{qu2020adversary,fischer2019online}) do not introduce overhead at either test or training time, whereas adversarial training~\cite{pinto2017robust,tessler2019action,tan2020robustifying} introduces an overhead only at training time. They are all more practical in this sense than reactive defenses because training can be done with a simulator, whereas reactive defenses require action at test time, often slowing down a system that has to provide answers in real time. In fact, the defense applied in this scenario \cite{wu2022crop,wang2025MultiAgentRobust} are all proactive. In addition, modifying RL systems that use previously proposed techniques~\cite{huang2019online} to predict the influence of other agents in the environment could increase the security of AD. 
Given that many autonomous systems still involve human supervision (i.e., SAE Level 2–3), explainability can help identify the presence of backdoor triggers or adversarial examples showing in a screen the regions on which the system is focusing \cite{sajid24}. Human-in-the-loop can be a useful resource to avoid dangerous situations, even though the explanation may be manipulated by the attack \cite{NEURIPS2019_bb836c01}. 

\subsection{Future Research Directions} 

The development of defenses for AD is still in its early stages, leaving many open research directions. Privacy remains a key challenge, as recent work shows that DRL models can be stolen by observing states and actions, yet no defense currently exists. Similarly, integrity violations lack effective solutions, since fine-tuning on clean data has proven insufficient. Availability violations, such as adversarial policy attacks, also remain undefended. 
Cooperative driving introduces significant challenges. The distributed nature of the system increases vulnerability, as some agents may be compromised, and even legitimate agents communicating over the network are exposed to data tampering, enabling evasion, poisoning, and privacy attacks. Furthermore, malicious agents may collude to pursue objectives that conflict with the overall system goals. However, there is serious lack of literature on security of CDS.
Future AD defenses should specify hardware requirements, quantify computational overhead, support system-level evaluation, and comply with industrial safety standards.
Finally, end-to-end and integrated pipelines for AD require special attention. Unlike modular architectures, where perception, planning, and control can be analyzed and secured or monitored independently, end-to-end systems tightly couple these components through shared representations and closed-loop decision making. 
As a result, small perturbations introduced at the perception level may propagate throughout the pipeline, leading to cascading vulnerabilities that manifest only at the control stage. 
Providing robustness across interconnected modules is, therefore, an open challenge. Future defenses should investigate sensor-fusion consistency checks that detect cross-modal inconsistencies before they affect planning decisions. 
Moreover, joint adversarial training across perception and control components may mitigate cross-stage error amplification, although its computational cost and stability remain open issues. 

\section{Attack and Defense Evaluation Framework}
A consistent evaluation framework is necessary to enable meaningful comparison across existing works on DRL security in AD. Building on the metrics discussed in Sections~\ref{sec:attack_metrics} and~\ref{sec:defense_metrics}, we summarize the key dimensions to jointly consider when assessing attacks and defenses.

\myparagraph{Attacks.} For attacks, evaluation should combine generic DRL metrics (\eg, Attack Success Rate, reward degradation, policy performance drop, perturbation magnitude, and transferability) with autonomous-driving-specific safety indicators such as collision rate, time-to-collision, crash frequency, and traffic rule violations. Beyond outcome-based measures, assessments should explicitly account for real-world constraints, including real-time perturbation feasibility, sensor noise, environmental variability, latency, and multi-agent traffic interactions.

\myparagraph{Defenses.} For defenses, evaluation must reflect their operational nature. Proactive methods should be assessed for robustness against adaptive attacks and the trade-off between robustness and nominal performance. Reactive mechanisms should additionally be evaluated based on detection accuracy, response latency, recovery effectiveness, and their impact on safety and traffic flow. 
In particular, in defenses based on MARL cooperation, the scalability issue should be analyzed and addressed.
Existing evaluation platforms such as CARLA~\cite{Dosovitskiy17} enable controlled experimentation and reproducibility and support safety-oriented metrics in complex traffic scenarios. However, their abstraction of sensing pipelines, simplified vehicle dynamics, and limited modeling of the physical world may constrain the realism of security evaluations, highlighting the need for complementary real-world testing and hardware-in-the-loop validation.

\section{Ethical, legal, and regulatory aspects}
Vehicles have an enormous potential to harm both passengers and pedestrians physically. 
In this section, we provide a broader perspective from our analysis and discuss related ethical problems as well as legal and regulatory implications.

\myparagraph{Ethical aspects.} Testing AD has already led to casualties.\footnote{\url{https://www.tesladeaths.com}} Even under benign circumstances, the differing behavior of AD when comparing to human drivers has led to resentment towards AD in, for example, San Francisco.\footnote{\url{https://www.traffictechnologytoday.com/videos/video-angry-mob-attacks-autonomous-vehicle-in-san-francisco.html}}compared From an ethical perspective, there is thus a strong duty to limit negative impacts from AD testing to the public, especially when testing adversaries. Implications include responsible experiments involving vehicles, responsible disclosure of reported attacks, and responsible treatment of program code to craft attacks. 

\myparagraph{Legislation and Regulation.} Regulations around vehicles are about, for example, general product functional safety (ISO-26262~\cite{ISO-26262}). However, these documents cover mainly traditional software and hardware components, and the application of these vehicle product safety regulations does not directly apply to AI~\cite{8860247}. 
2Other standards define levels of driving automation (SAE J3016~\cite{SAEJ3016:2021}) or cover safety of autonomous products, including vehicles (UL 4600~\cite{UL4600:2020}).
These standards do not inspect the internal vulnerabilities of learning-based decision-making components and therefore neglect a significant portion of the vulnerabilities related to DRL that we outlined in this study.
One example is that AI models depend on their training data and are thus variable. Also, such safety standards do not entail an attacker, a shortcoming that limits their usefulness when considering AI security as discussed in this survey.\footnote{\url{https://oecd.ai/en/wonk/when-it-comes-to-ai-incidents-safety-and-security-are-not-the-same}} 
To address security concerns, automotive regulations have been extended to include cybersecurity risk management, most notably through ISO/SAE 21434~\cite{ISO-21434}, and more recently ISO/TS 5083~\cite{SO-TS5083}, which targets the cybersecurity of automated driving systems.
Yet, these standards provide essential guidance for identifying threats, but they fail to address AI-specific security concerns~\cite{2024arXiv241214855B}. 
More recently, ISO/PAS 8800~\cite{ISO-PAS-8800} has been introduced as the first standard to explicitly extend safety engineering practices to AI/ML components in vehicles, covering aspects such as data quality, model validation, and AI-specific failure modes. However, this standard is mostly oriented toward unintentional AI failures, e.g., distributional shift or insufficient training data, rather than adversarially induced ones, and does not incorporate a threat model comparable to the one we propose.
These shortcomings have been acknowledged and are taken into account in recent legislation, such as the European AI Act~\cite{act2024eu}. The corresponding standards, also for AI security, are currently under development\footnote{\url{https://www.cencenelec.eu/areas-of-work/cen-cenelec-topics/artificial-intelligence/}}. 
This survey complements existing standards by reviewing DRL-specific attack surfaces and defense mechanisms, providing relevant insights to support the implementation and evolution of secure AD systems.

\section{Conclusion}
\label{sec:concluding-remark}

In this survey, we categorize existing attacks and defenses on DRL. Unlike previous works, we provide a framework that helps system designers identify suitable defenses for specific threats. Since AD is a critical application, we examined it in greater depth, revealing that research on securing DRL-based AD systems still requires further study. We also highlight which ML attacks have yet to be adapted to RL and which lack defenses, offering both practitioners and researchers insight into potential threats and promising future directions.


\section{Acknowledgement}
This work has been partly supported by Huawei Technologies Co. Ltd., project ``Deep Reinforcement Learning Key Security Technologies'' (Agreement No: TC20201118006).



\bibliographystyle{IEEEtran}


\begin{IEEEbiography}
[{\includegraphics[width=1in,height=1.25in,clip,keepaspectratio]{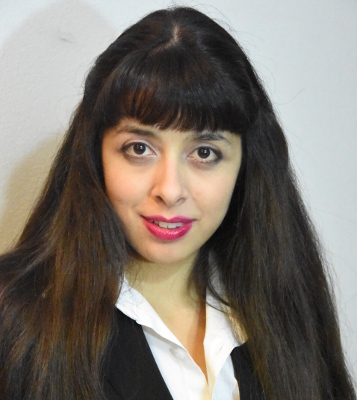}}]
{Ambra Demontis} is an Assistant Professor at the University of Cagliari, Italy. She received her M.Sc. degree (Hons.) in Computer Science and her Ph.D. degree in Electronic Engineering and Computer Science, respectively, in 2014 and 2018. Her research focuses on secure machine learning. She is an Associate Editor for  Pattern Recognition and TIFS. She is a Member of the IEEE and the IAPR.
\end{IEEEbiography}

\begin{IEEEbiography}
[{\includegraphics[width=1in,height=1.25in,clip,keepaspectratio]{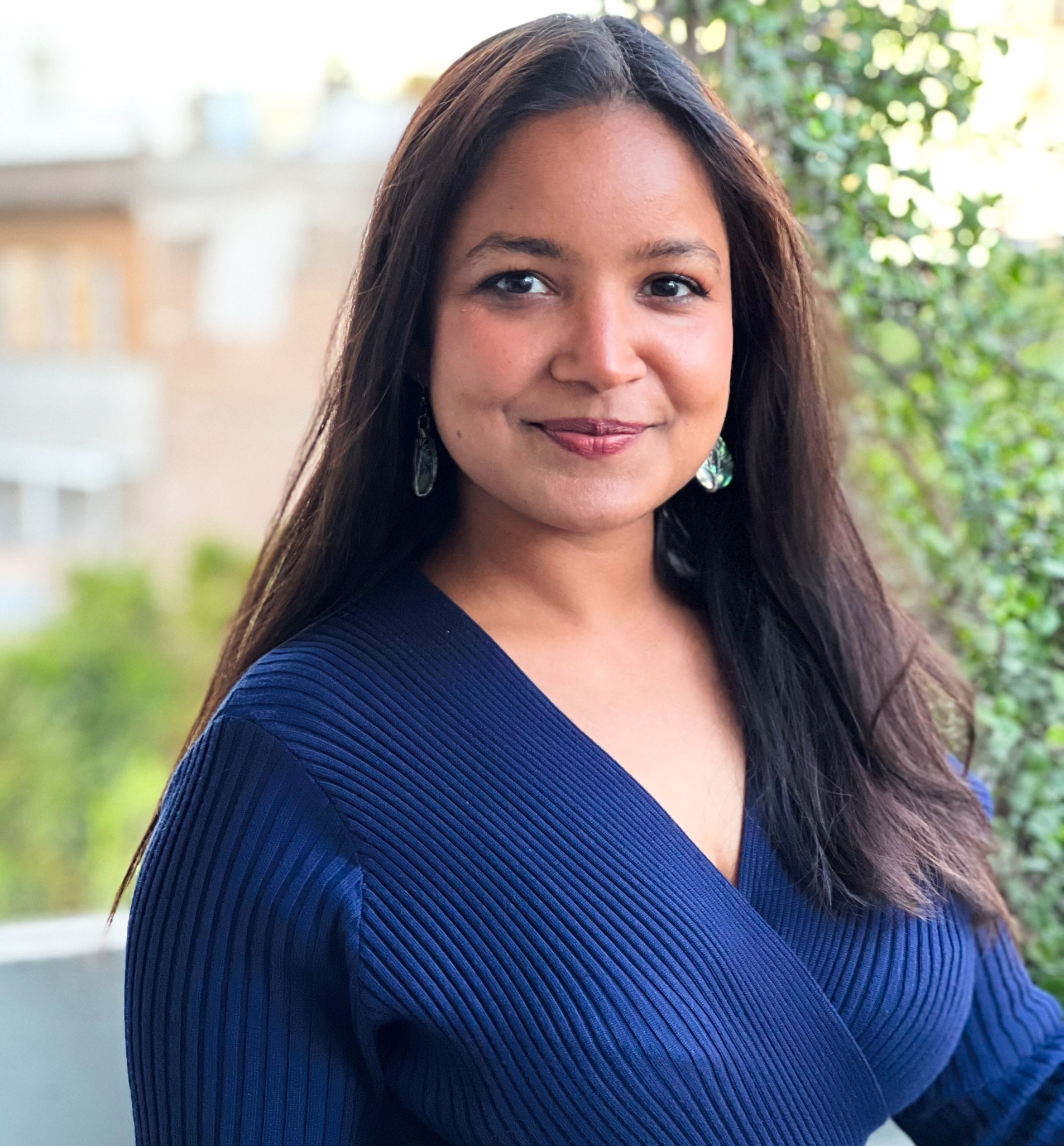}}]
{Srishti Gupta} is a PhD in student in the Italian National Ph.D. program in AI at Sapienza University, Rome co-hosted by the University of Cagliari, Italy. She received her M.Sc. from the University of Arizona, US and B.Tech from Bharati Vidyapeeth College in Delhi, India. Her research interests include Continual Learning, Out-of-Distribution Detection, and the security of LLM models. She serves as a reviewer for several journals and conferences.
\end{IEEEbiography}
\begin{IEEEbiography}
[{\includegraphics[width=1in,height=1.25in,clip,keepaspectratio]{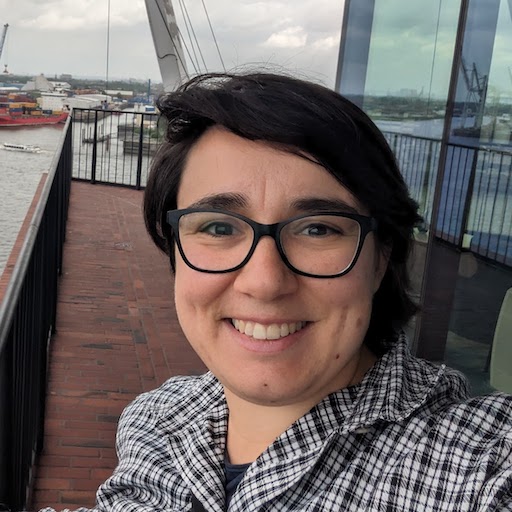}}]{Maura Pintor} is an Assistant Professor at the PRA Lab, in the Department of Electrical and Electronic Engineering of the University of Cagliari, Italy. She received her PhD in Electronic and Computer Engineering from the University of Cagliari in 2022. Her research focuses on machine learning security.
\end{IEEEbiography}
\begin{IEEEbiography}
[{\includegraphics[width=1in,height=1.25in,clip,keepaspectratio]{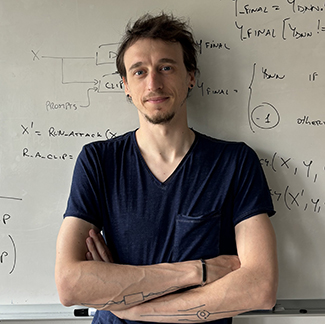}}]
{Luca Demetrio} is a Tenure-Track Assistant Professor at the University of Genova (Italy), where he also received his Ph.D. in 2021. He is currently studying the security of Windows malware detectors implemented with ML techniques, as well as new techniques to improve the evaluation of ML models using tools.
\end{IEEEbiography}
\begin{IEEEbiography}
[{\includegraphics[width=1in,height=1.25in,clip,keepaspectratio]{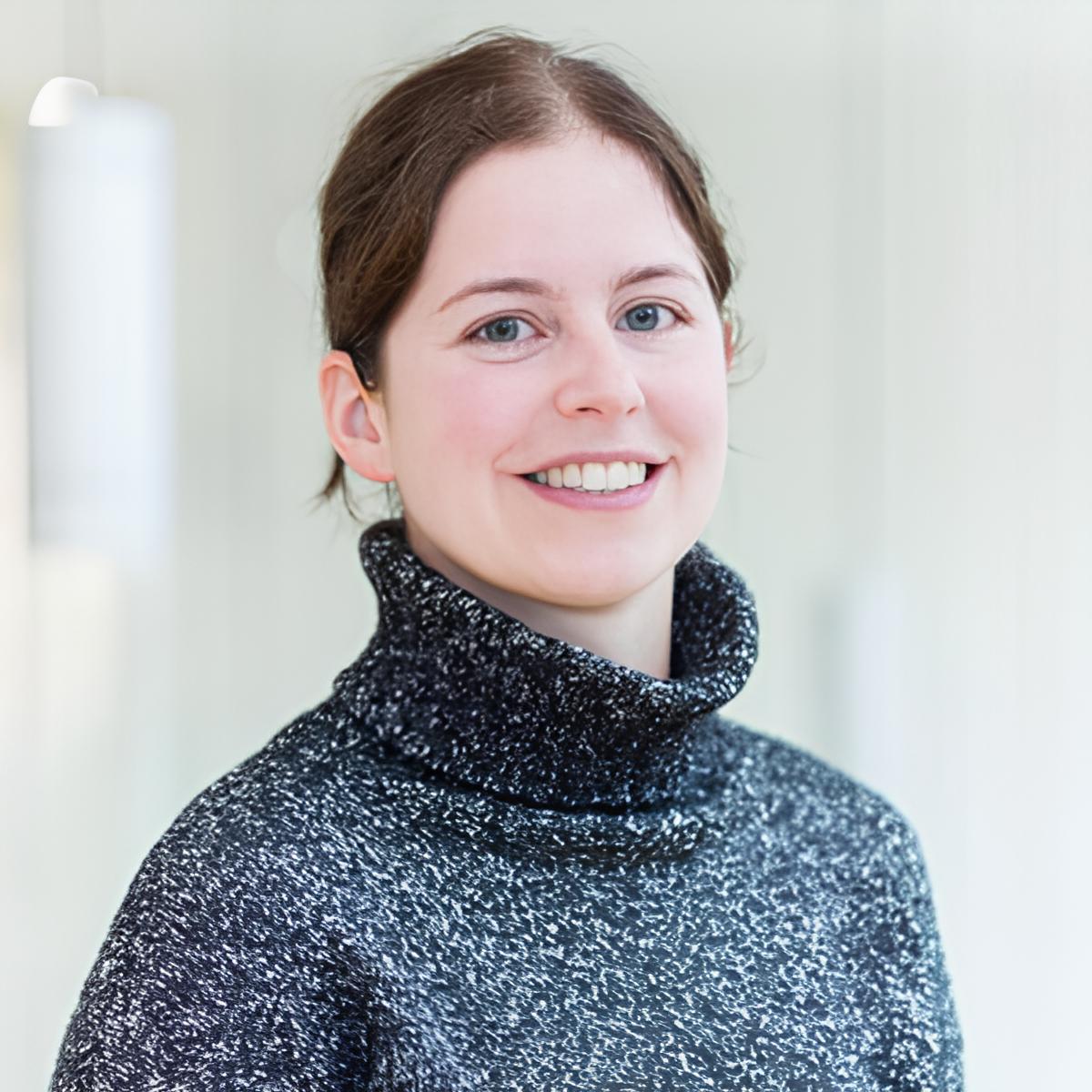}}]
{Kathrin Grosse} is an Independent Researcher. Her work bridges the gap between AI security research and industry needs. She earned her Ph.D. at CISPA Helmholtz Center, Saarland University, in 2021, followed by postdoctoral research with Battista Biggio in Cagliari, Italy, and Alexandre Alahi at EPFL, Switzerland. Since 2025, she has been a member of the European research network ELLIS. She contributes to ETSI standards for AI incident reporting.
\end{IEEEbiography}
\begin{IEEEbiography}
[{\includegraphics[width=1in,height=1.25in,clip,keepaspectratio]{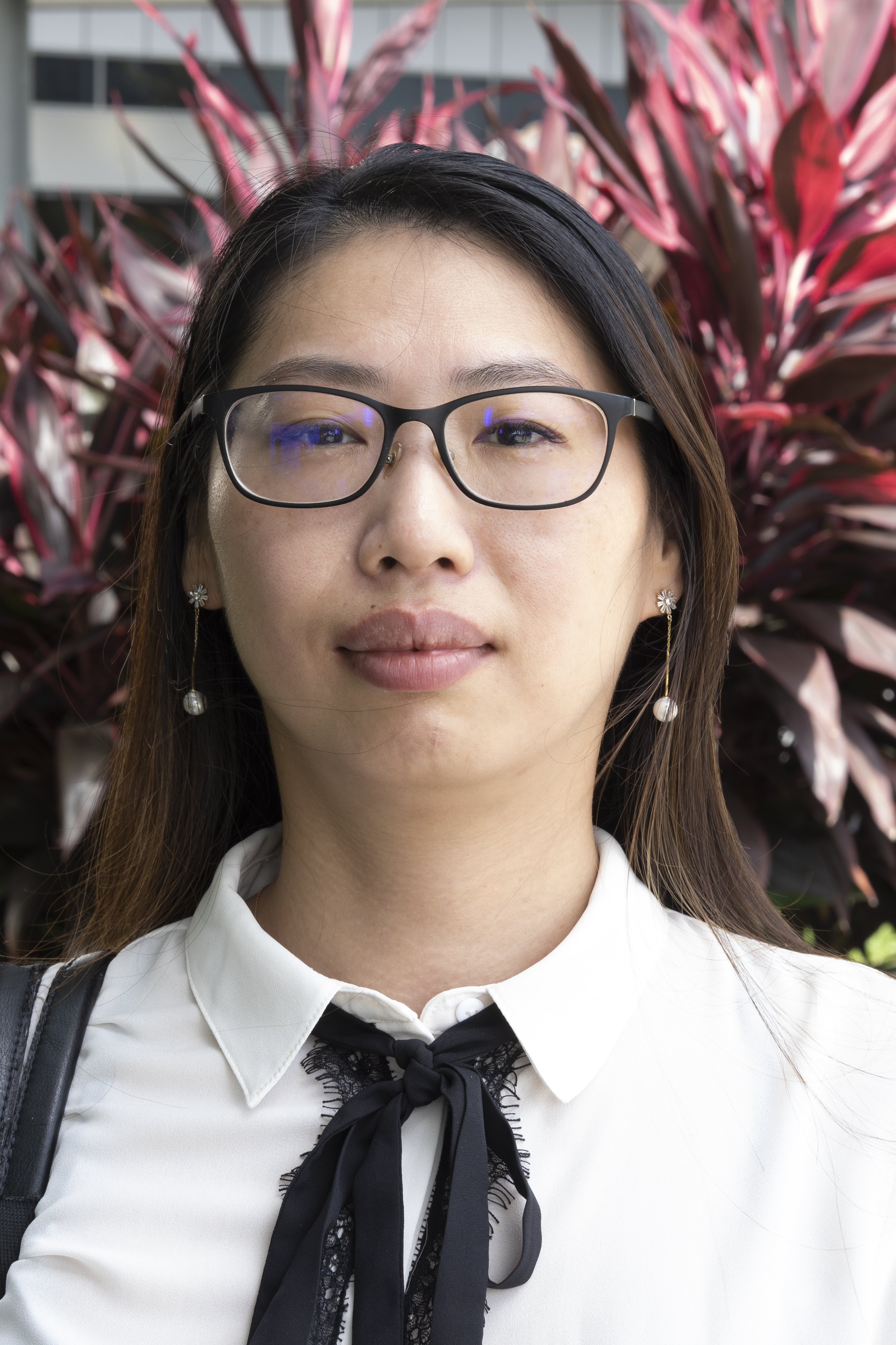}}]
{Hsiao-Ying Lin} is a principal researcher in Shield Labs at Huawei Technologies France. Her research interests include adversarial machine learning, applied cryptography, and security issues in automotive areas. She received the MS and PhD degrees in computer science from National Chiao Tung University, Taiwan, in 2005 and 2010, respectively.
\end{IEEEbiography}
\begin{IEEEbiography}
[{\includegraphics[width=1in,height=1.25in,clip,keepaspectratio]{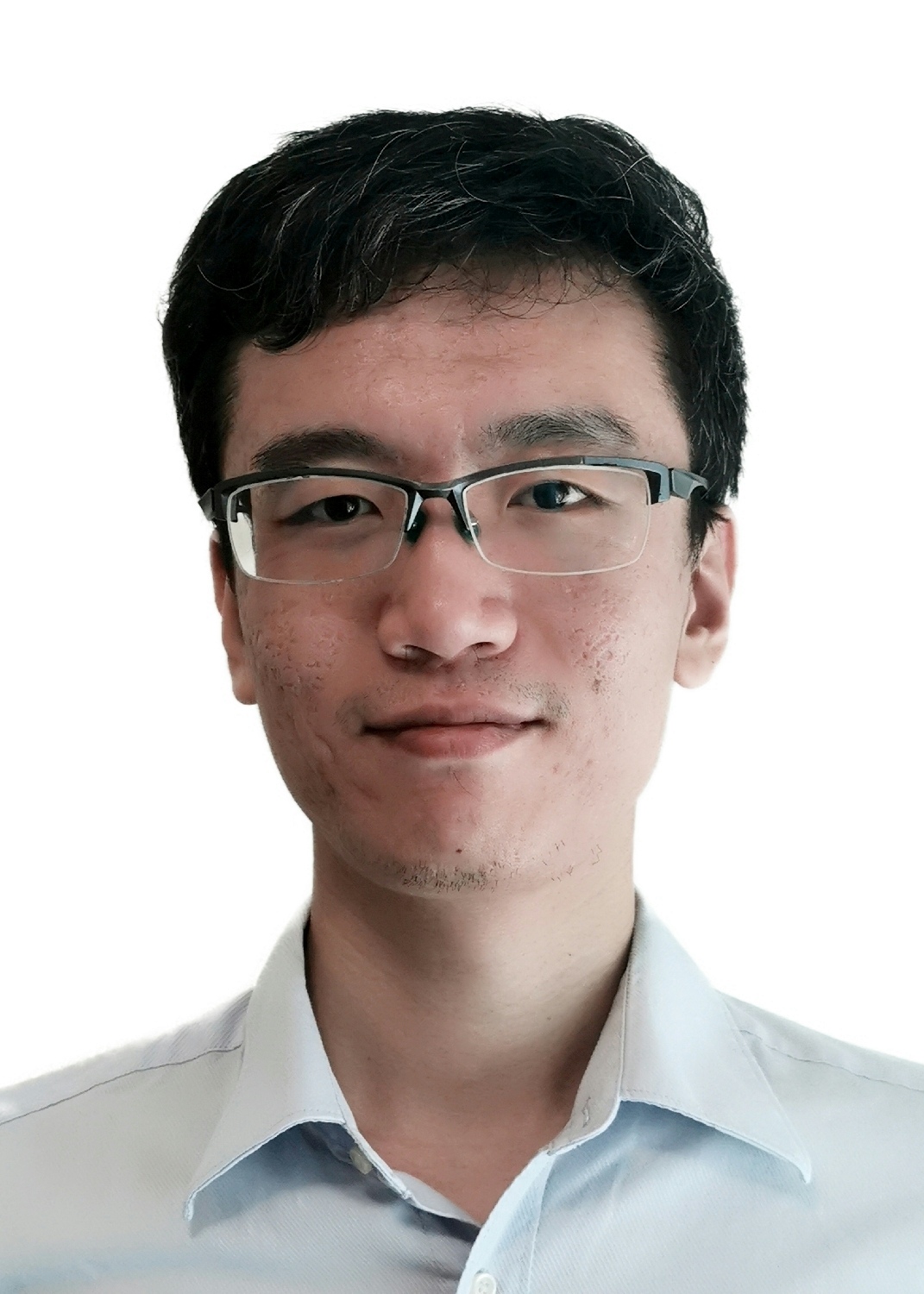}}]
{Chengfang Fang} obtained both his bachelor's and Ph.D. from the National University of Singapore with the Tata Consultancy Award and the Research Achievement Award. He joined Huawei and continued to work on security and privacy protection across domains, such as ML, internet of things, mobile device and biometrics. He has been working on these areas for over a decade, publishing over 30 research papers and obtaining over 20 patents in the domain. He is currently a principal researcher at Huawei Singapore Research Center.
\end{IEEEbiography}

\begin{IEEEbiography}
[{\includegraphics[width=1in,height=1.25in,clip,keepaspectratio]{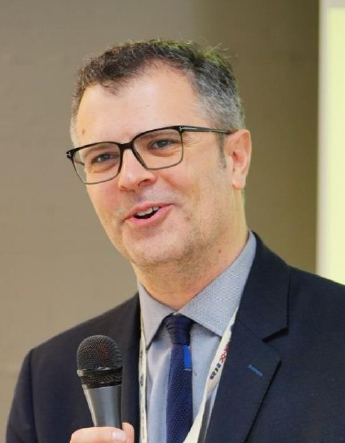}}]{Battista Biggio} is a Full Professor of Computer Engineering at the University of Cagliari, Italy. He has provided pioneering contributions to machine learning security. His paper ``Poisoning Attacks against Support Vector Machines'' won the 2022 ICML Test of Time Award.  He chaired IAPR TC1 (2016-2020) and served as AE for IEEE TNNLS and IEEE CIM. He is the Associate EiC for Pattern Recognition and serves as an AC for NeurIPS and the IEEE Symp. S\&P. He is an IEEE and AAIA Fellow, ACM Senior Member, and a member of IAPR, AAAI, and ELLIS.
\end{IEEEbiography}

\begin{IEEEbiography}
[{\includegraphics[width=1in,height=1.25in,clip,keepaspectratio]{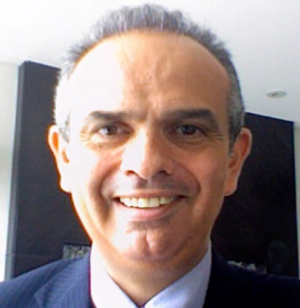}}] {Fabio Roli} is a Full Professor of Computer Engineering at the University of Genova, Italy. He has been appointed Fellow of the IEEE and Fellow of the International Association for Pattern Recognition. He is a recipient of the Pierre Devijver Award for his contributions to statistical pattern recognition.
\end{IEEEbiography}

\end{document}